\documentclass[journal,10pt,letterpaper]{IEEEtran}
\usepackage{cite}
\usepackage{graphicx}
\usepackage{amsmath,amssymb}
\usepackage[hyphens]{url}
\usepackage{xcolor}
\usepackage{array}
\usepackage{booktabs}
\usepackage{multirow}
\usepackage{placeins}
\newcolumntype{C}[1]{>{\centering\arraybackslash}p{#1}}
\newcommand{\incre}[1]{\textcolor{green!50!black}{#1}}
\newcommand{\decre}[1]{\textcolor{red!70!black}{#1}}
\long\def\cn#1{}

\title{Unleashing the Potential of Vision-Language Models for Generalizable AI-Generated Image Detection}
\author{Weihan Cai, Hao Tan, Zichang Tan, Jun Wan, and Xinping Gao%
\thanks{Corresponding author: Jun Wan.}%
\thanks{Weihan Cai, Hao Tan, and Jun Wan are with the State Key Laboratory of Multimodal Artificial Intelligence Systems, Institute of Automation, Chinese Academy of Sciences, Beijing 100190, China (e-mail: \mbox{caiweihan2025@ia.ac.cn}; \mbox{tanhao2023@ia.ac.cn}; \mbox{jun.wan@ia.ac.cn}).}%
\thanks{Zichang Tan is with Sangfor Technologies Inc., Shenzhen 518055, China (e-mail: \mbox{tanzichang@foxmail.com}).}%
\thanks{Xinping Gao is with Purple Mountain Laboratories, Nanjing 211111, China (e-mail: \mbox{gaoxinping@pmlabs.com.cn}).}}
\begin{document}

\maketitle

\begin{abstract}
A linear classifier trained on frozen DINOv3 features can outperform specialized detectors of AI-generated images in the wild. This success has motivated efforts to further adapt DINOv3 for detection, but we find that the same evaluation protocol underestimates the potential of a vision-language model. When linear probes are trained on images from the same generator, Perception Encoder (PE) performs worse than DINOv3 on in-the-wild benchmarks. Yet real and generated images form more distinct local groups in PE’s frozen feature space. Image–text pretraining gives PE a source of supervision directly relevant to this distinction: captions can identify an image as AI-generated, allowing PE to learn high-level semantics about image provenance. Conventional linear probing uses only the image encoder. We introduce Semantic Prototype Calibration (SPC), which instead initializes the linear head of the frozen image encoder with text embeddings that describe real and AI-generated images. Across nine in-the-wild subsets, SPC raises mean accuracy from 84.8\% to 92.9\% for PE and from 81.4\% to 98.5\% for MoEViE. Both models also improve on controlled benchmarks. MoEViE-SPC outperforms specialized detectors with more elaborate task-specific designs. These results establish SPC as a simple and effective baseline for generalizable AI-generated image detection.
\end{abstract}
\cn{摘要
在 DINOv3 的冻结特征上训练一个线性分类器，就能在野外 AI 生成图像检测中超过专用检测器。这一结果推动了针对 DINOv3 的进一步适配，但我们发现，同样的评测方式低估了视觉-语言模型的检测潜力。当线性探针使用来自同一生成器的图像进行训练时，Perception Encoder（PE）在野外基准上的表现不及 DINOv3。然而，直接考察这些野外图像的冻结特征，真实图像与生成图像在 PE 的特征空间中形成了更清晰的局部分组。图文预训练为 PE 提供了与这一区分直接相关的监督：配文可以明确指出一张图像由 AI 生成，使 PE 能够学到有关图像来源的高层语义。常规线性探针却只使用图像编码器。为此，我们提出语义原型校准（SPC），用描述真实图像和 AI 生成图像的文本嵌入初始化冻结图像编码器的线性头。在九个野外子集上，SPC 将 PE 的平均准确率从 84.8\% 提升至 92.9\%，将 MoEViE 从 81.4\% 提升至 98.5\%。两种模型在控制基准上的整体表现也有所提升。MoEViE-SPC 超过了采用更复杂任务专用设计的方法。这些结果为可泛化的 AI 生成图像检测确立了一个简单有效的基线。
}

\section{Introduction}
\cn{1. 引言}

\IEEEPARstart{D}{etecting} AI-generated images in the wild requires a detector to recognize images from generators it has never seen, often after compression or online sharing.
A texture or frequency pattern that reliably distinguishes real and generated images in the training set may no longer be informative under these conditions.
One promising approach is to use visual representations learned on a broader range of images before detector training.
\emph{Simplicity Prevails}~\cite{SimplicityPrevails} showed that even a linear classifier trained on frozen features from a modern vision foundation model can outperform specialized detectors in the wild.
DINOv3~\cite{DINOv3} was particularly effective under this protocol, establishing a strong baseline for generalizable AI-generated image (AIGI) detection.
Several subsequent methods have built on DINOv3 to further improve detection performance~\cite{FGTS,MIRROR,ARA}.
\cn{
在野外场景中，检测器需要识别来自未知生成器的图像，而这些图像往往还经历过压缩或网络传播。
在训练集中能够可靠区分真假图像的纹理或频率模式，到了这些条件下可能就不再有效。
一种有前景的做法，是利用视觉模型在检测训练之前从更广泛的图像中学到的表征。
Simplicity Prevails [SimplicityPrevails] 表明，即使只在现代视觉基础模型的冻结特征上训练一个线性分类器，也能在野外基准上超过专用检测器。
DINOv3 [DINOv3] 在这一设置下表现尤为突出，为可泛化的 AI 生成图像（AIGI）检测建立了一个有竞争力的基线。
后续多项工作在 DINOv3 的基础上进一步提升了检测性能 [FGTS, MIRROR, ARA]。
}

\begin{figure}[t]
\centering
\includegraphics[width=\columnwidth]{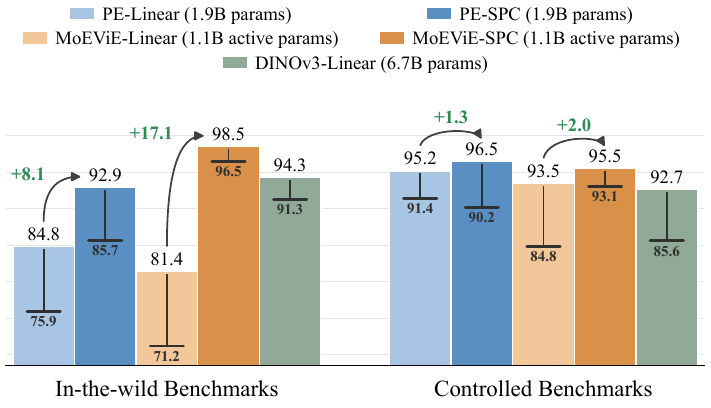}
\caption{Both PE and MoEViE benefit from text initialization, with larger gains in the wild than on controlled benchmarks.
Bars show the mean and horizontal markers show the minimum within each group. Green labels give the mean gain over the corresponding linear probe.
Across the nine in-the-wild subsets, including those from SynthWildX, MoEViE-SPC reaches a mean of 98.5\% and a minimum of 96.5\%.}
\cn{用文本嵌入初始化线性头后，PE 和 MoEViE 均有提升，且两者在野外基准上的增益都大于在控制基准上的增益。
柱高和横线分别表示各组内的均值和最低值，绿色数字表示相对同一编码器线性探针的平均增益。
在包含 SynthWildX 在内的九个野外子集上，MoEViE-SPC 的平均平衡准确率达到 98.5\%，最低值也达到 96.5\%。
}
\label{fig:spc-improvement}
\end{figure}

\begin{figure*}[t]
\centering
\includegraphics[width=\textwidth]{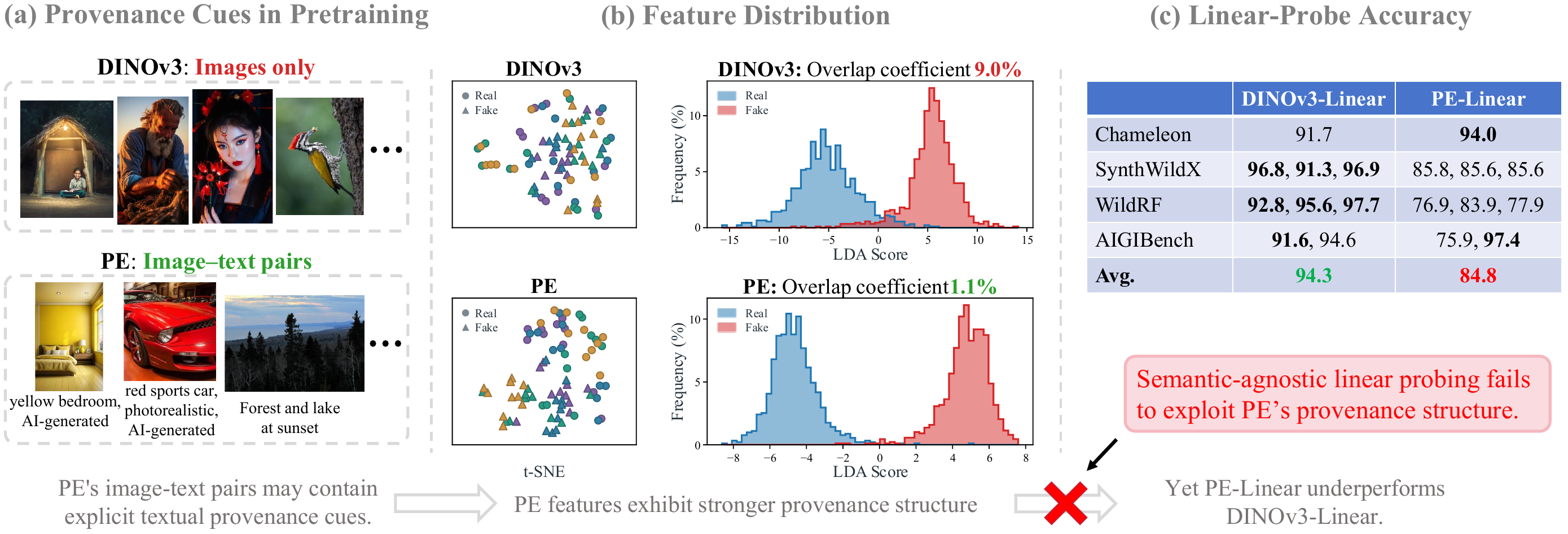}
\caption{PE's frozen features separate real and generated images more clearly than DINOv3's, yet its linear probe performs worse in the wild.
(a) DINOv3 is pretrained on images alone, whereas PE learns from image--text pairs whose captions can explicitly identify an image as AI-generated.
(b) For features pooled from in-the-wild datasets, t-SNE (left) shows clearer local class separation for PE, and the real and generated LDA score distributions (right) overlap less.
(c) When both probes are trained on GenImage SD-v1.4, PE-Linear achieves a mean balanced accuracy of 84.8\% across nine in-the-wild subsets, including those from SynthWildX, compared with 94.3\% for DINOv3-Linear.}
\cn{PE 的冻结特征比 DINOv3 的冻结特征更清楚地区分了真实图像与生成图像，但 PE 的线性探针在野外基准上的表现却更差。
(a) DINOv3 仅使用图像进行预训练，而 PE 从图文对中学习，配对文本可以明确指出图像由 AI 生成。
(b) 汇总野外数据集的特征后，t-SNE 可视化（左）显示两类图像在 PE 特征空间中的局部分离更清晰，两类特征的 LDA 得分分布（右）也更少重叠。
(c) 两个探针均在 GenImage SD-v1.4 上训练时，PE-Linear 在包含 SynthWildX 在内的九个野外子集上的平均平衡准确率为 84.8\%，低于 DINOv3-Linear 的 94.3\%。
}
\label{fig:feature-distribution}
\end{figure*}

Vision--language models offer a different starting point for this task.
During image--text pretraining, a caption can identify an image as AI-generated without naming the generator or describing any particular visual artifact.
Perception Encoder (PE)~\cite{PE} can thus learn a high-level concept of image provenance: images that differ in content, style, or generator can still be associated with the same description of how they were produced.
Conventional linear probing, however, uses only PE's image encoder.
The distinction between real and generated images must then be learned by a randomly initialized classifier from the available detection data.
A low probe score therefore leaves open an important question: do the frozen features contain too little information to distinguish the two classes, or does the classifier fail to use information already present in them?
\cn{
视觉-语言模型为这项任务提供了另一种起点。
在图文预训练中，配文可以明确指出一张图像由 AI 生成，而无须指明具体的生成器，也无须描述某种特定的视觉痕迹。
Perception Encoder（PE）[PE] 由此能够学到关于图像来源的高层语义：即使图像的内容、风格和生成器各不相同，它们仍然可以与同一种来源描述相对应。
然而，常规线性探针只使用 PE 的图像编码器，真实图像与生成图像之间的区别，需要由随机初始化的分类器从有限的检测数据中学得。
因此，较低的探针得分留下了一个值得区分的问题：究竟是冻结特征中缺少区分两类图像的信息，还是这些信息已经存在，却没有被分类器充分利用？
}

To investigate these possibilities, we compare PE and DINOv3 on the same in-the-wild images, examining both their linear-probe performance and the structure of their frozen features (Fig.~\ref{fig:feature-distribution}).
When both probes are trained on the same GenImage SD-v1.4 data, PE achieves a mean balanced accuracy of 84.8\% across nine in-the-wild subsets, compared with 94.3\% for DINOv3.
Judged by probe performance alone, DINOv3 would be the natural choice.
Yet the frozen features reveal a different pattern.
Real and generated images form more distinct local groups in PE's feature space, and the two classes overlap less when projected onto a discriminant direction.
This discrepancy suggests that conventional linear probing underestimates PE's potential for in-the-wild detection.
\cn{
为考察这两种可能，我们在同一批野外图像上比较 PE 和 DINOv3，既考察线性探针的检测性能，也直接观察冻结特征的分布（图~\ref{fig:feature-distribution}）。
当两个探针均在相同的 GenImage SD-v1.4 数据上训练时，PE 在九个野外子集上的平均平衡准确率为 84.8\%，而 DINOv3 达到 94.3\%。
仅从探针的表现来看，DINOv3 是更自然的选择。
然而，冻结特征呈现出了不同的结果：真实图像与生成图像在 PE 的特征空间中形成了更清晰的局部分组，两类特征沿判别方向投影后的分布也更少重叠。
这一反差表明，常规线性探针低估了 PE 在野外检测中的潜力。
}

This observation motivates a simple change to how the classifier learns from PE's features.
In a linear head, each class weight defines a direction along which an image feature is scored.
Although a conventional probe learns these directions from a random initialization, PE's text encoder already provides a way to specify an initial direction for each class.
It maps descriptions such as ``a real photo'' and ``AI generated'' into the same space as the image features.
We therefore use the corresponding text embeddings to initialize the two class weights and refine them on labeled images while keeping the image encoder frozen.
We call this approach Semantic Prototype Calibration (SPC).
It preserves the simplicity of linear probing while allowing the classifier to build on the notion of image provenance learned during pretraining.
\cn{
这一观察促使我们重新考虑分类器利用 PE 特征的方式。
在线性分类头中，每个类别权重都定义了一个方向，线性头沿该方向对图像特征进行评分。
常规线性探针从随机初始化开始学习这些方向，但 PE 的文本编码器本身就可以为每个类别提供一个初始方向：它能够将“a real photo”和“AI generated”等描述映射到图像特征所在的空间。
因此，我们用对应的文本嵌入初始化两个类别权重，再利用带标签的图像对其进行校准，同时保持图像编码器冻结。
我们将这一方法称为语义原型校准（SPC）。
这样既保留了线性探针的简洁性，也使分类器能够从预训练阶段学到的图像来源概念出发进行学习。
}

This change yields its largest gains in the wild (Fig.~\ref{fig:spc-improvement}).
Across the nine in-the-wild subsets, SPC raises PE's mean from 84.8\% to 92.9\%.
The same initialization raises the mean for MoEViE~\cite{MoEViE} from 81.4\% to 98.5\%, with a score of at least 96.5\% on every subset.
Both encoders also improve on the controlled benchmarks.
With the architecture of a linear probe, MoEViE-SPC thus achieves a higher in-the-wild average than every specialized detector in our comparison, including methods with considerably more elaborate designs.
These results establish SPC as a simple and effective baseline against which future AIGI detectors can be evaluated.
\cn{
这一改动在野外基准上带来的增益尤为明显（图~\ref{fig:spc-improvement}）。
在九个野外子集上，SPC 将 PE 的均值从 84.8\% 提升至 92.9\%。
同样的初始化方式将 MoEViE [MoEViE] 的均值从 81.4\% 提升至 98.5\%，即使在表现最弱的子集上也达到 96.5\%。
两种编码器在控制基准上的整体表现同样有所提升。
MoEViE-SPC 保留了线性探针的结构，却在野外基准上超过了所有参评的专用检测器，包括设计更为复杂的方法。
这些结果为后续 AIGI 检测研究确立了一个简单有效的比较基线。
}

Our main contributions are as follows.
\cn{本文的主要贡献如下。}
\begin{itemize}
    \item We find that conventional linear probing underestimates PE's potential for in-the-wild AIGI detection. Although PE scores lower than DINOv3 under linear probing, its frozen features show clearer local separation between real and AI-generated images.
    \cn{我们发现，常规线性探针低估了 PE 在野外场景中的 AIGI 检测潜力。尽管 PE 在线性探针下的表现不及 DINOv3，真实图像与 AI 生成图像在 PE 冻结特征空间中的局部分离却更为清晰。}
    \item To better realize this potential, we introduce Semantic Prototype Calibration (SPC). Simply initializing the linear head with text embeddings that describe real and AI-generated images raises PE's average across nine in-the-wild subsets from 84.8\% to 92.9\%.
    \cn{为了更充分地发掘这一潜力，我们提出语义原型校准（SPC）。仅用描述真实图像和 AI 生成图像的文本嵌入初始化线性头，SPC 就将 PE 在九个野外子集上的均值从 84.8\% 提升至 92.9\%。}
    \item SPC provides a simple yet effective baseline for AIGI detection. The same change raises MoEViE's average to 98.5\% (+17.1\%) across the nine in-the-wild subsets, surpassing all specialized detectors in our comparison, including those with more elaborate task-specific designs.
    \cn{SPC 为 AIGI 检测提供了一个简单有效的新基线。同样的改动也将 MoEViE 在九个野外子集上的均值提升至 98.5\%（+17.1\%），超过了参与比较的所有专用检测器，包括采用更复杂的任务专用设计的方法。}
\end{itemize}

\section{Related Work}
\cn{2. 相关工作}

\noindent \textbf{Foundation models for AIGI detection.}
UnivFD~\cite{UnivFD} showed that a linear classifier trained on frozen CLIP image features can generalize across generators.
Subsequent work explored task-specific architectures and training strategies for adapting pretrained visual features to AIGI detection~\cite{AIDE,DDA,Effort}.
More recently, studies of modern vision foundation models~\cite{SimplicityPrevails,VFM} found that a linear probe alone can outperform specialized detectors in the wild.
DINOv3~\cite{DINOv3} emerged as a strong baseline and has since been used in several detection methods~\cite{FGTS,ARA}.
These findings motivate a closer look at how a linear probe uses the information available in a frozen encoder.
\cn{用于 AIGI 检测的基础模型
UnivFD [UnivFD] 表明，在冻结的 CLIP 图像特征上训练线性分类器，即可实现跨生成器泛化。
后续研究通过专用结构和训练策略，将预训练视觉特征用于 AIGI 检测 [AIDE, DDA, Effort]。
近期针对现代视觉基础模型的研究发现，仅训练线性探针就能在野外评测中超过专用检测器 [SimplicityPrevails, VFM]。
DINOv3 [DINOv3] 由此成为一个强基线，后续多种检测方法也以其为基础 [FGTS, ARA]。
这些结果促使我们进一步考察，线性探针如何利用冻结编码器中已有的信息。
}

\begin{figure*}[t]
\centering
\includegraphics[width=\textwidth]{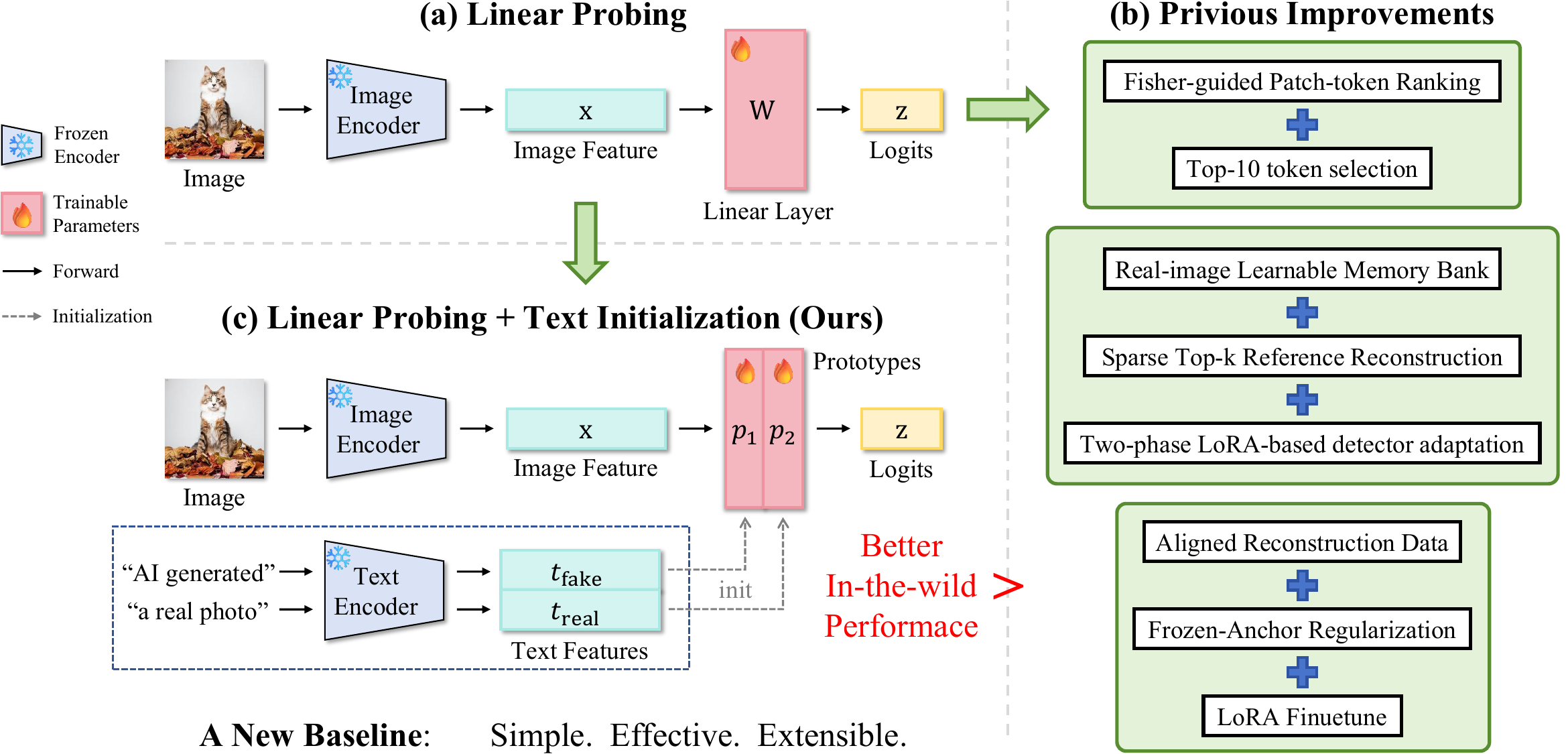}
\caption{Semantic Prototype Calibration (SPC) preserves the architecture of a linear probe and changes how its class weights are initialized.
(a) Conventional linear probing trains a randomly initialized binary head on features from a frozen image encoder.
(b) Prior detectors introduce task-specific components.
(c) SPC initializes the two class weights with text embeddings of ``AI generated'' and ``a real photo,'' then calibrates them on labeled images. The dashed arrows denote initialization.
With this change, MoEViE-SPC ranks first overall on the in-the-wild benchmarks.}
\cn{语义原型校准（SPC）保留线性探针的结构，仅改变类别权重的初始化方式。
(a) 常规线性探针在冻结图像编码器提取的特征上训练随机初始化的二分类头。
(b) 已有检测器引入了针对检测任务的设计。
(c) SPC 用“AI generated”和“a real photo”的文本嵌入分别初始化两个类别权重，再用带标签的图像进行校准。虚线箭头表示初始化。
仅通过这一改动，MoEViE-SPC 的野外综合表现就在参评方法中排名第一。
}
\label{fig:architecture}
\end{figure*}

\noindent \textbf{Vision-language models for AIGI detection.}
Vision-language models offer an additional source of information for detection.
Image--text pretraining places images and textual descriptions in a shared embedding space~\cite{CLIP}, where descriptions of how an image was produced can serve as references for its provenance.
Previous CLIP-based detectors have used prompts or text prototypes to guide the learning of forensic representations~\cite{C2PCLIP,MiraGe,GCSNet}.
For a detector built on a frozen encoder, this shared space raises a simpler question. Can a description of each class directly specify how the linear head scores an image?
\cn{用于 AIGI 检测的视觉-语言模型
视觉-语言模型为检测提供了额外的信息来源。
图文预训练将图像与文本描述映射到共享的嵌入空间 [CLIP]，因此，描述图像产生方式的文本可以作为判断其来源的参照。
已有基于 CLIP 的检测方法利用提示词或文本原型引导取证表征的学习 [C2PCLIP, MiraGe, GCSNet]。
对于建立在冻结编码器上的检测器，这一共享空间引出了一个更简单的问题。能否直接用每个类别的文本描述，给出线性头对图像进行评分的依据？
}

\noindent \textbf{A geometric view of the linear classification head.}
A linear classification head assigns each class a weight vector.
From a geometric perspective, this vector defines a direction along which the classifier measures the image feature.
Work on face recognition~\cite{SphereFace,NormFace,CosFace} makes this interpretation explicit. When the feature and class weight are normalized, their dot product is their cosine similarity.
In a vision-language model, a class description can supply such a direction.
CLIP~\cite{CLIP} uses text embeddings as the weights of a zero-shot classifier.
We adopt this perspective for AIGI detection. The two weights of the binary head serve as prototypes for real and AI-generated images.
SPC initializes each prototype with the corresponding text embedding and calibrates the two directions with labels.
\cn{线性分类头的几何视角
线性分类头为每个类别学习一个权重向量。从几何上看，这个向量定义了一个方向，分类器沿该方向对图像特征进行评分。
人脸识别领域的研究清楚地体现了这一点 [SphereFace, NormFace, CosFace]。当图像特征与类别权重均经过归一化时，二者的点积就是余弦相似度。
在视觉-语言模型中，类别的文本描述可以给出这样的方向。
CLIP [CLIP] 将文本嵌入作为零样本分类器的权重。
我们从这一视角理解 AIGI 检测中的二分类头。它的两个类别权重分别是真实图像和 AI 生成图像的原型。
SPC 用对应的文本嵌入初始化这两个原型，再利用标签校准这两个方向。
}

\section{Method}
\cn{3. 方法}

We first examine how clearly frozen image features distinguish real from generated images and how well a conventional linear probe uses this information.
We then interpret the weights of the binary head as class prototypes and introduce Semantic Prototype Calibration (SPC), which initializes these prototypes with text embeddings before fitting the head to labeled images.
\cn{
我们首先考察冻结的图像特征能在多大程度上区分真实图像与生成图像，以及常规线性探针能否充分利用这些信息。
随后，我们将二分类头的权重视为类别原型，并引出语义原型校准（SPC）。该方法用文本嵌入初始化两个原型，再用带标签的图像训练分类头。
}

\subsection{Provenance Semantics Under Conventional Linear Probing}
\cn{3.1 常规线性探针对来源语义的利用}

Exposure to real and generated images during visual pretraining can equip image-only encoders such as DINOv3~\cite{DINOv3} with visual cues for AIGI detection~\cite{SimplicityPrevails,VFM}.
PE~\cite{PE} has an additional source of supervision. As Figure~\ref{fig:feature-distribution}(a) illustrates, a caption can explicitly identify an image as AI-generated, linking its visual features to its provenance.
In PE's shared embedding space, this link can be interpreted by comparing image features with textual concepts such as ``AI generated''~\cite{SimplicityPrevails}.
Figure~\ref{fig:feature-distribution}(b) compares frozen features pooled from in-the-wild datasets. Both encoders show local separation between real and generated images, but the separation is clearer for PE. The overlap coefficient between the two classes' LDA score distributions is 1.1\% for PE and 9.0\% for DINOv3.
\cn{
通过在视觉预训练中接触真实图像和生成图像，DINOv3 [DINOv3] 等纯视觉编码器就能学到有助于 AIGI 检测的视觉线索 [SimplicityPrevails, VFM]。
PE [PE] 还具有额外的监督来源。如图~\ref{fig:feature-distribution}(a) 所示，文本可能明确指出一幅图像由 AI 生成，从而将其视觉特征与图像来源联系起来。
在 PE 的共享嵌入空间中，可以通过比较图像特征与“AI generated”等文本概念来解释这种联系 [SimplicityPrevails]。
图~\ref{fig:feature-distribution}(b) 比较了在野外数据集上汇总的冻结特征。两种编码器的特征都呈现出真实图像与生成图像的局部分离，其中 PE 的分离更清晰。两类图像的 LDA 得分分布在 PE 上的重叠系数为 1.1\%，在 DINOv3 上为 9.0\%。
}

Under conventional linear probing, however, the ranking reverses. When both probes are trained on GenImage SD-v1.4~\cite{GenImage}, PE-Linear averages 84.8\% across the nine in-the-wild subsets, compared with 94.3\% for DINOv3-Linear (Figure~\ref{fig:feature-distribution}(c)).
Thus, the clearer class separation observed in PE's frozen features does not translate directly into better cross-generator detection.

In this setting, the linear head is randomly initialized and learns its class weights from a single training source~\cite{UnivFD,SimplicityPrevails}. The text encoder provides no direct guidance during this process.
This observation motivates a closer look at the class weights themselves and how their initial directions could reflect the provenance semantics learned during pretraining.
\cn{
然而，采用常规线性探针时，两种编码器的表现顺序恰好相反。两个探针均在 GenImage SD-v1.4 [GenImage] 上训练后，PE-Linear 在九个野外子集上的均值为 84.8\%，低于 DINOv3-Linear 的 94.3\%（图~\ref{fig:feature-distribution}(c)）。
可见，PE 冻结特征中更清晰的类别分离并未直接转化为更好的跨生成器检测性能。

在这一设置下，线性头从随机初始化出发，仅依靠单一训练源学习类别权重 [UnivFD, SimplicityPrevails]，文本编码器并未直接参与这一过程。
因此，我们进一步考察分类头的类别权重，以及如何让这些权重的初始方向体现预训练中学到的来源语义。
}

\begin{table*}[!t]
\caption{Balanced accuracy (\%) on in-the-wild benchmarks. Avg. is the unweighted mean across the nine subsets. For SPC, parentheses give the change in Avg. from the corresponding Linear baseline. Bold indicates the highest accuracy in each column.}
\cn{野外基准上的平衡准确率（\%）。Avg. 为九个子集的等权平均值。SPC 均值后的括号表示相对同一编码器 Linear 基线的差值。粗体表示各准确率列的最高值。}
\label{tab:wild}
\centering \small
\setlength{\tabcolsep}{2mm}
\resizebox{\textwidth}{!}{%
\begin{tabular}{l|c|ccc|ccc|cc|c}
\toprule
\multirow{2}{*}{\textbf{Method}} &
\multirow{2}{*}{\textbf{Chameleon}} &
\multicolumn{3}{c|}{\textbf{SynthWildX}} &
\multicolumn{3}{c|}{\textbf{WildRF}} &
\multicolumn{2}{c|}{\textbf{AIGIBench}} &
\multirow{2}{*}{\textbf{Avg.}} \\
\cmidrule(lr){3-5}\cmidrule(lr){6-8}\cmidrule(lr){9-10}
& & \textbf{DALL-E 3} & \textbf{Firefly} & \textbf{Midjourney} & \textbf{Facebook} & \textbf{Reddit} & \textbf{Twitter} & \textbf{SocialRF} & \textbf{CommunityAI} & \\
\midrule
\multicolumn{11}{c}{\textit{Traditional AIGI Detectors}} \\
\midrule
NPR~\cite{NPR} & 59.9 & 43.6 & 61.3 & 44.5 & 78.1 & 61.0 & 51.3 & 53.3 & 55.1 & 56.5 \\
AIDE~\cite{AIDE} & 63.1 & 63.4 & 48.8 & 51.9 & 57.8 & 71.5 & 45.8 & 57.3 & 53.7 & 57.0 \\
DDA~\cite{DDA} & 83.5 & 91.1 & 84.7 & 91.6 & 85.3 & 82.5 & 89.3 & 79.9 & 88.9 & 86.3 \\
\midrule
\multicolumn{11}{c}{\textit{Specialized AIGI Detectors with Modern VFMs}} \\
\midrule
FGTS~\cite{FGTS} & 91.9 & 97.2 & 92.3 & 96.8 & 95.9 & 96.3 & 98.1 & 95.0 & 95.6 & 95.5 \\
MIRROR~\cite{MIRROR} & 90.7 & 95.9 & 88.4 & 94.9 & 97.1 & 96.6 & 96.4 & 87.6 & 93.4 & 93.4 \\
ARA~\cite{ARA} & 95.4 & 97.8 & 95.2 & 97.6 & 97.8 & 98.3 & 98.5 & \textbf{96.8} & 97.0 & 97.2 \\
\midrule
\multicolumn{11}{c}{\textit{Frozen VFM + Linear Probing}} \\
\midrule
OpenCLIP-Linear & 70.5 & 57.9 & 60.1 & 55.6 & 72.2 & 60.9 & 54.3 & 54.1 & 66.4 & 61.3 \\
SigLIP2-Linear & 87.2 & 84.1 & 78.4 & 81.8 & 77.5 & 86.2 & 88.9 & 77.3 & 87.5 & 83.2 \\
MetaCLIP2-Linear & 93.4 & 73.5 & 72.6 & 73.3 & 80.3 & 79.1 & 80.6 & 68.9 & 92.3 & 79.3 \\
PE-Linear & 94.0 & 85.8 & 85.6 & 85.6 & 76.9 & 83.9 & 77.9 & 75.9 & 97.4 & 84.8 \\
DINOv3-Linear & 91.7 & 96.8 & 91.3 & 96.9 & 92.8 & 95.6 & 97.7 & 91.6 & 94.6 & 94.3 \\
MoEViE-Linear & \textbf{98.0} & 77.4 & 76.9 & 77.3 & 75.6 & 85.6 & 72.7 & 71.2 & 98.0 & 81.4 \\
\midrule
\multicolumn{11}{c}{\textit{Frozen VFM + SPC (Ours)}} \\
\midrule
PE-SPC (Ours) & 96.1 & 92.8 & 92.5 & 92.5 & 95.6 & 90.6 & 93.1 & 85.7 & 97.1 & 92.9 \incre{(+8.1)} \\
MoEViE-SPC (Ours) & 97.6 & \textbf{98.4} & \textbf{97.8} & \textbf{98.3} & \textbf{100.0} & \textbf{99.1} & \textbf{99.7} & 96.5 & \textbf{99.0} & \textbf{98.5} \incre{(+17.1)} \\
\bottomrule
\end{tabular}
}
\end{table*}

\subsection{Semantic Prototype Calibration}
\label{sec:spc}
\cn{3.2 语义原型校准}

Let $\mathbf{x}=f_I(I)\in\mathbb{R}^{1\times n}$ denote the feature extracted from image $I$ by the frozen image encoder $f_I$.
A $K$-class linear head has a weight matrix $W\in\mathbb{R}^{n\times K}$ and a bias vector $\mathbf{b}$.
Writing the $k$-th column of $W$ as $p_k^\top$, we can express its output:
\begin{equation}
\begin{aligned}
\mathbf{z}
&= \mathbf{x}W+\mathbf{b}\\
&= \mathbf{x}\left[p_1^\top,\ldots,p_K^\top\right]+\mathbf{b}\\
&= \left[\mathbf{x}p_1^\top,\ldots,\mathbf{x}p_K^\top\right]+\mathbf{b}\\
&= \left[\langle\mathbf{x},p_1\rangle,\ldots,\langle\mathbf{x},p_K\rangle\right]+\mathbf{b}.
\end{aligned}
\end{equation}
Each inner product compares the image feature with a reference vector for one class. We therefore interpret $p_1,\ldots,p_K$ as learnable class prototypes in the image embedding space.
\cn{
记 x = f_I(I) 为冻结图像编码器 f_I 从图像 I 中提取的 n 维行向量。对于 K 类线性分类头，权重矩阵 W 的形状为 n×K，偏置向量为 b。将 W 的第 k 列记为 p_kᵀ，分类头的输出可以写为

z = xW + b
  = x [p₁ᵀ, …, p_Kᵀ] + b
  = [x p₁ᵀ, …, x p_Kᵀ] + b
  = [⟨x, p₁⟩, …, ⟨x, p_K⟩] + b

每个内积都将图像特征与一个类别的参考向量进行比较。因此，我们可以将 p₁ 到 p_K 理解为图像嵌入空间中一组可学习的类别原型。
}

In a vision-language model, images and their textual descriptions share an embedding space~\cite{CLIP}. A description of each class can therefore provide an initial direction for its prototype.
We use the text encoder $f_T$ to embed the prompts ``AI generated'' and ``a real photo'' and initialize the two columns of the detection head,
\begin{equation}
\begin{aligned}
p_{\mathrm{fake}}^{(0)} &= \mathcal{N}(f_T(\mbox{``AI generated''})),\\
p_{\mathrm{real}}^{(0)} &= \mathcal{N}(f_T(\mbox{``a real photo''})),\\
W^{(0)} &= \left[(p_{\mathrm{fake}}^{(0)})^\top,(p_{\mathrm{real}}^{(0)})^\top\right],
\end{aligned}
\end{equation}
where $\mathcal{N}$ denotes $\ell_2$ normalization.
\cn{
在视觉-语言模型中，图像与文本描述位于共享的嵌入空间 [CLIP]，因此每个类别的文本描述都可以为相应的原型提供初始方向。
我们使用文本编码器 f_T 分别编码“AI generated”和“a real photo”，用它们初始化检测头的两列权重。

p_fake⁽⁰⁾ = N(f_T("AI generated"))
p_real⁽⁰⁾ = N(f_T("a real photo"))
W⁽⁰⁾ = [(p_fake⁽⁰⁾)ᵀ, (p_real⁽⁰⁾)ᵀ]

其中，N 表示 ℓ₂ 归一化。
}

Starting from $W^{(0)}$, we train the linear head on labeled real and generated images while keeping the image encoder frozen.
We call this step \emph{prototype calibration}. Before training begins, the two prototypes already encode the distinction between ``AI generated'' and ``a real photo'' learned during image--text pretraining.
These initial directions can reduce the head's reliance on cues specific to the training source, such as textures associated with a single generator.
\cn{
从 W⁽⁰⁾ 出发，我们使用带标签的真实图像和生成图像训练线性头，图像编码器始终保持冻结。
我们将这一步称为原型校准。在检测训练开始前，两个原型就已带有图文预训练所学到的语义，分别指向“AI generated”和“a real photo”。
以这两个方向为起点进行校准，有助于减少分类头对训练源特有线索的依赖，例如某一生成器特有的纹理。
}

As Figure~\ref{fig:architecture} shows, SPC is simpler to implement than approaches that introduce patch-token selection or adapt the image encoder~\cite{FGTS,ARA}.
The text encoder computes the initial weights once. At inference, each image passes through the same frozen image encoder and binary linear head as in linear probing, without additional parameters or computation.
\cn{
如图~\ref{fig:architecture} 所示，SPC 相比引入图像块 token 筛选或调整图像编码器的改进方案，实现更为简单 [FGTS, ARA]。
文本编码器只在初始化时计算一次权重。推理时，图像仍然只经过冻结的图像编码器和二分类线性头，与线性探针相比，模型参数量和计算量均无增加。
}

\section{Results}
\cn{4. 结果}

\subsection{Experimental Setup}
\cn{4.1 实验设置}

\noindent \textbf{Benchmarks.}
We train our detectors on the Stable Diffusion v1.4 training split of GenImage~\cite{GenImage} and evaluate generalization on two groups of benchmarks.
The in-the-wild group comprises Chameleon~\cite{AIDE}, SynthWildX~\cite{SynthWildX}, WildRF~\cite{WildRF}, and the SocialRF and CommunityAI subsets of AIGIBench~\cite{AIGIBench}.
Table~\ref{tab:wild} reports the mean across their nine test subsets, weighting each subset equally.
The controlled group comprises GenImage, UnivFakeDetect~\cite{UnivFD}, AIGI-Now~\cite{AIGINow}, and AIGI-Holmes~\cite{AIGIHolmes}.
We evaluate all eight GenImage test splits, and follow the evaluation protocol of MIRROR~\cite{MIRROR} on UnivFakeDetect.
AIGI-Now and AIGI-Holmes test transfer to recent generators.
For AIGI-Now, we follow the nine-generator protocol, which includes GPT-4o.
The controlled average gives equal weight to the four benchmark means.
Throughout the paper, accuracy denotes balanced accuracy, the mean of the real-class and generated-class recalls.
\cn{评估基准
我们在 GenImage [GenImage] 的 Stable Diffusion v1.4 训练划分上训练检测器，并在野外和控制两组基准上评估泛化性能。
野外基准包括 Chameleon [AIDE]、SynthWildX [SynthWildX]、WildRF [WildRF]，以及 AIGIBench [AIGIBench] 中的 SocialRF 和 CommunityAI。
表~\ref{tab:wild} 对这九个测试子集等权取平均。
控制基准包括 GenImage、UnivFakeDetect [UnivFD]、AIGI-Now [AIGINow] 和 AIGI-Holmes [AIGIHolmes]。
我们评估 GenImage 的全部八个测试划分，并沿用 MIRROR [MIRROR] 的评测协议在 UnivFakeDetect 上进行测试。
AIGI-Now 和 AIGI-Holmes 用于检验模型向近期生成器的迁移能力。
对于 AIGI-Now，我们沿用包含 GPT-4o 的九生成器评测协议。
控制基准的总体得分为四个基准各自均值的等权平均。
全文中的准确率均指平衡准确率，即真实类召回率与生成类召回率的平均值。
}

\noindent \textbf{Baselines.}
Our main comparison pairs SPC with conventional linear probing on the same frozen encoder.
We evaluate both settings on PE~\cite{PE} and MoEViE~\cite{MoEViE}, and extend the comparison to OpenCLIP~\cite{OpenCLIP}, SigLIP2~\cite{SigLIP2}, and MetaCLIP2~\cite{MetaCLIP2} to examine how the effect of SPC varies across VLMs.
DINOv3-Linear~\cite{DINOv3} serves as our vision-only baseline following \emph{Simplicity Prevails}~\cite{SimplicityPrevails}.
We use the largest publicly available checkpoint in each model family.
We also compare with traditional AIGI detectors~\cite{NPR,AIDE,DDA} and specialized detectors built on modern VFMs~\cite{FGTS,MIRROR,ARA}.
\cn{比较方法
我们的主要对比是在同一冻结编码器上分别采用 SPC 和常规线性探针。
我们在 PE [PE] 和 MoEViE [MoEViE] 上评估这两种设置，并将比较扩展至 OpenCLIP [OpenCLIP]、SigLIP2 [SigLIP2] 和 MetaCLIP2 [MetaCLIP2]，以考察 SPC 的效果如何随视觉-语言模型而变化。
我们沿用 Simplicity Prevails [SimplicityPrevails] 的设置，以 DINOv3-Linear [DINOv3] 作为纯视觉基线。
各模型系列均使用公开可用的最大规模版本。
此外，我们还与传统 AIGI 检测器 [NPR, AIDE, DDA] 以及基于现代视觉基础模型的专用检测器 [FGTS, MIRROR, ARA] 进行比较。
}

\noindent \textbf{Implementation details.}
For both Linear and SPC, we freeze the encoder and train only the binary linear head for two epochs using AdamW with a batch size of 128.
We resize and center-crop each image to the input resolution of its encoder and use no additional data augmentation.
For Linear, we randomly initialize the head and use a learning rate of $1\times10^{-3}$, following \emph{Simplicity Prevails}~\cite{SimplicityPrevails}.
For SPC, we initialize the class weights with the text embeddings described in Section~\ref{sec:spc} and use the learning rate of $2\times10^{-5}$ selected on GenImage (Figure~\ref{fig:learning-rate-sensitivity}(a)).
To stabilize optimization during the early stage of prototype calibration, we initialize the biases for the generated and real classes to 1 and 0, respectively.
\cn{实现细节
对于 Linear 和 SPC，我们均冻结编码器，仅训练二分类线性头。两者都使用 AdamW 优化器，批大小为 128，训练两个 epoch。
输入图像经过缩放和中心裁剪，以匹配各编码器的输入分辨率，不使用额外的数据增强。
对于 Linear，我们沿用 Simplicity Prevails [SimplicityPrevails] 的设置，随机初始化线性头，学习率设为 1e-3。
对于 SPC，我们使用第 3.2 节所述的文本嵌入初始化类别权重，并采用在 GenImage 上选出的学习率 2e-5（图~\ref{fig:learning-rate-sensitivity}(a)）。
为稳定原型校准早期阶段的优化过程，我们将生成类和真实类的偏置分别初始化为 1 和 0。
}

\begin{table*}[!t]
\caption{Mean balanced accuracy (\%) on four controlled benchmarks. Each benchmark column reports its subset mean. Avg. is the unweighted mean of the four benchmark scores and is reported only for methods with complete results. For SPC, parentheses give the change in Avg. from the corresponding Linear baseline. A dash indicates an unavailable benchmark result. Bold indicates the highest accuracy in each column.}
\cn{四个控制基准上的平均平衡准确率（\%）。每个基准列给出该基准各子集的均值。Avg. 为四个基准得分的等权平均值，仅对具有完整结果的方法报告。SPC 均值后的括号表示相对同一编码器 Linear 基线的差值。横线表示该基准的结果缺失，粗体表示各准确率列的最高值。}
\label{tab:controlled-summary}
\centering \small
\setlength{\tabcolsep}{2.5mm}
\resizebox{0.8\textwidth}{!}{%
\begin{tabular}{l|cccc|c}
\toprule
\textbf{Method} & \textbf{GenImage} & \textbf{UnivFakeDetect} & \textbf{AIGI-Now} & \textbf{AIGI-Holmes} & \textbf{Avg.} \\
\midrule
\multicolumn{6}{c}{\textit{Traditional AIGI Detectors}} \\
\midrule
NPR~\cite{NPR} & 51.5 & 51.1 & 61.9 & 79.4 & 61.0 \\
AIDE~\cite{AIDE} & 61.2 & 77.9 & 67.2 & 97.0 & 75.8 \\
DDA~\cite{DDA} & 91.7 & 69.4 & 70.1 & 96.4 & 81.9 \\
\midrule
\multicolumn{6}{c}{\textit{Specialized AIGI Detectors with Modern VFMs}} \\
\midrule
FGTS~\cite{FGTS} & 92.6 & 89.4 & 83.4 & 96.0 & 90.4 \\
MIRROR~\cite{MIRROR} & -- & 88.2 & -- & -- & -- \\
ARA~\cite{ARA} & -- & -- & 87.1 & 98.2 & -- \\
\midrule
\multicolumn{6}{c}{\textit{Frozen VFM + Linear Probing}} \\
\midrule
OpenCLIP-Linear & 76.8 & 63.7 & 86.3 & 81.8 & 77.2 \\
SigLIP2-Linear & 93.9 & 86.4 & 84.1 & 96.9 & 90.3 \\
MetaCLIP2-Linear & 90.6 & 84.3 & 91.4 & 93.4 & 89.9 \\
PE-Linear & 97.2 & 91.4 & 94.6 & 97.6 & 95.2 \\
DINOv3-Linear & 96.7 & 91.1 & 85.6 & 97.4 & 92.7 \\
MoEViE-Linear & 95.4 & 84.8 & 94.2 & 99.4 & 93.5 \\
\midrule
\multicolumn{6}{c}{\textit{Frozen VFM + SPC (Ours)}} \\
\midrule
PE-SPC (Ours) & \textbf{98.6} & 90.2 & \textbf{98.0} & 99.2 & \textbf{96.5} \incre{(+1.3)} \\
MoEViE-SPC (Ours) & 96.1 & \textbf{93.2} & 93.1 & \textbf{99.5} & 95.5 \incre{(+2.0)} \\
\bottomrule
\end{tabular}
}
\end{table*}

\subsection{Generalization to In-the-Wild Data}
\cn{4.2 野外基准上的泛化性能}

Initializing the linear head with text embeddings substantially improves in-the-wild generalization for both encoders (Table~\ref{tab:wild}).
PE's mean rises from 84.8\% to 92.9\%, and MoEViE's rises from 81.4\% to 98.5\%.
The improvement is especially clear on SocialRF, MoEViE's weakest subset under both settings, where its score increases from 71.2\% to 96.5\%.
With the same inference architecture as a conventional linear probe, MoEViE-SPC achieves a higher mean than every specialized detector in the comparison.
These results support using SPC as a baseline for in-the-wild AIGI detection.
\cn{
如表~\ref{tab:wild} 所示，用文本嵌入初始化线性头，大幅提升了两种编码器在野外基准上的泛化性能。
PE 的均值从 84.8\% 提升至 92.9\%，MoEViE 则从 81.4\% 提升至 98.5\%。
SocialRF 在两种设置下都是 MoEViE 表现最弱的子集，但其得分从 71.2\% 提升到了 96.5\%。
在保持常规线性探针推理结构的情况下，MoEViE-SPC 的均值超过了所有参与比较的专用检测器。
这些结果支持将 SPC 作为野外 AIGI 检测的新基线。
}

\subsection{Performance on Controlled Benchmarks}
\cn{4.3 控制基准上的性能}

SPC also improves average performance on the four controlled benchmarks, where conventional linear probing already performs well (Table~\ref{tab:controlled-summary}).
The four-benchmark mean rises from 95.2\% to 96.5\% for PE and from 93.5\% to 95.5\% for MoEViE.
PE-SPC achieves the highest mean among methods with results on all four benchmarks.
Tables in appendix provide the results for individual subsets.
\cn{
在四个控制基准上，常规线性探针已经表现良好，SPC 进一步提升了两种编码器的平均性能（表~\ref{tab:controlled-summary}）。
PE 的均值从 95.2\% 提升至 96.5\%，MoEViE 则从 93.5\% 提升至 95.5\%。
在具有完整四项结果的方法中，PE-SPC 的均值最高。各子集的详细结果见附录中的表格。
}

\section{Analysis}

\subsection{Is Prototype Calibration Necessary?}
\cn{是否需要原型校准？}

\begin{table}[!t]
\caption{Mean balanced accuracy (\%) before and after prototype calibration. NoTrain uses the initial text prototypes directly, SPC calibrates them with labeled images, and Linear denotes standard linear probing. For each encoder and benchmark group, $\Delta_{\mathrm{NoTrain}}$ is the SPC mean minus the NoTrain mean. Bold marks the highest mean in each benchmark group.}
\cn{原型校准前后的平均平衡准确率（\%）。NoTrain 直接使用初始文本原型，SPC 使用带标签的图像对其进行校准，Linear 表示常规线性探针。对于每个编码器和每组基准，Δ_NoTrain 均为 SPC 均值减去 NoTrain 均值。粗体表示在各组基准上的最高均值。}
\label{tab:no-training-ablation}
\centering\small
\setlength{\tabcolsep}{2.5pt}
\resizebox{0.8\columnwidth}{!}{%
\begin{tabular}{l|l|cc|cc}
\toprule
\multirow{2}{*}{\textbf{Model}} & \multirow{2}{*}{\textbf{Setting}} & \multicolumn{2}{c|}{\textbf{In-the-Wild}} & \multicolumn{2}{c}{\textbf{Controlled}} \\
\cmidrule(lr){3-4}\cmidrule(lr){5-6}
 & & \textbf{Avg.} & \textbf{$\Delta_{\mathrm{NoTrain}}$} & \textbf{Avg.} & \textbf{$\Delta_{\mathrm{NoTrain}}$} \\
\midrule
\multirow{3}{*}{PE} & Linear & 84.8 & -- & 95.2 & -- \\
 & NoTrain & 86.5 & -- & 93.9 & -- \\
 & SPC & \textbf{92.9} & \incre{\textbf{+6.4}} & \textbf{96.5} & \incre{\textbf{+2.6}} \\
\midrule
\multirow{3}{*}{MoEViE} & Linear & 81.4 & -- & 93.5 & -- \\
 & NoTrain & 91.6 & -- & 95.4 & -- \\
 & SPC & \textbf{98.5} & \incre{\textbf{+6.9}} & \textbf{95.5} & \incre{\textbf{+0.1}} \\
\bottomrule
\end{tabular}
}
\end{table}

We examine how much of SPC's detection performance is already present in the initial text prototypes and how much is gained through supervised calibration.
In the NoTrain setting, we perform no detector training and classify images directly using the text embeddings of ``AI generated'' and ``a real photo'' as class prototypes.
As shown in Table~\ref{tab:no-training-ablation}, NoTrain outperforms Linear in the wild for both PE (86.5\% vs.\ 84.8\%) and MoEViE (91.6\% vs.\ 81.4\%).
These results suggest that the pretrained alignment between images and descriptions of their origin already provides a useful signal for detection.
\cn{
本节考察初始文本原型本身具备怎样的检测能力，以及有监督校准能在此基础上带来多大提升。
在 NoTrain 设置下，我们不进行任何检测器训练，直接使用“AI generated”和“a real photo”的文本嵌入作为类别原型进行分类。
如表~\ref{tab:no-training-ablation} 所示，在野外基准上，PE 的 NoTrain 得分为 86.5\%，高于 Linear 的 84.8\%，MoEViE 的 NoTrain 也以 91.6\% 超过 Linear 的 81.4\%。
这些结果表明，预训练中形成的图像与来源描述之间的对齐关系，已经为检测提供了有用的线索。
}

Calibrating the same prototypes raises in-the-wild accuracy to 92.9\% for PE and 98.5\% for MoEViE, gains of 6.4\% and 6.9\% over NoTrain.
On the controlled benchmarks, calibration improves PE by 2.6\% and MoEViE by 0.1\%.
The larger gains in the wild suggest that supervised calibration helps adapt the initial decision boundary to the detection task, even when the text prototypes already distinguish real from generated images reasonably well.
\cn{
对同一组原型进行校准后，PE 和 MoEViE 的野外准确率分别达到 92.9\% 和 98.5\%，较 NoTrain 提升 6.4\% 和 6.9\%。
在控制基准上，校准使 PE 和 MoEViE 分别提升 2.6\% 和 0.1\%。
野外基准上更明显的增益表明，即使初始文本原型已经具备较好的真假判别能力，利用带标签的样本校准原型，仍能使决策边界更适合检测任务。
}

\subsection{Learning-Rate Sensitivity}
\cn{5.3 学习率敏感性}

\begin{figure}[t]
\centering
\includegraphics[width=\columnwidth]{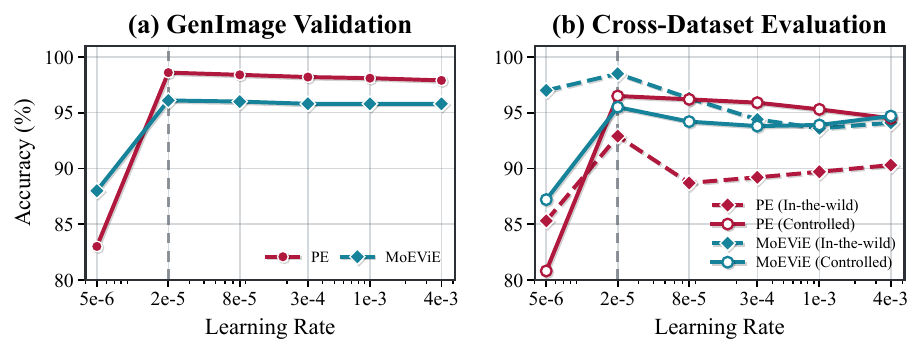}
\caption{Learning-rate selection on GenImage and evaluation across benchmarks. (a) GenImage validation performance of PE-SPC and MoEViE-SPC at six learning rates. (b) Mean performance on the in-the-wild benchmarks (dashed lines, diamonds) and controlled benchmarks (solid lines, circles). The vertical dashed line in each panel marks $2\times10^{-5}$, the rate selected on GenImage for both encoders.}
\cn{在 GenImage 上选择学习率并在各组基准上评估其表现。(a) PE-SPC 和 MoEViE-SPC 在六个学习率下的 GenImage 验证结果。(b) 两种模型在野外基准（菱形虚线）和控制基准（圆点实线）上的均值。两幅图中的竖直虚线均标出两种编码器在 GenImage 上选出的学习率 2e-5。}
\label{fig:learning-rate-sensitivity}
\end{figure}

We select the learning rate for SPC on GenImage. With all other training settings fixed, we evaluate six rates from $5\times10^{-6}$ to $4\times10^{-3}$.
Both PE-SPC and MoEViE-SPC achieve their best GenImage validation result at $2\times10^{-5}$ (Figure~\ref{fig:learning-rate-sensitivity}(a)). We therefore use this rate for both encoders across benchmarks.
On the in-the-wild benchmarks, the rate selected on GenImage also yields the highest mean among the six rates tested for both models (Figure~\ref{fig:learning-rate-sensitivity}(b)).
\cn{
我们在 GenImage 上为 SPC 选择学习率。在其余训练设置相同的条件下，我们测试了从 5e-6 到 4e-3 的六个学习率。
如图~\ref{fig:learning-rate-sensitivity}(a) 所示，PE-SPC 和 MoEViE-SPC 均在 2e-5 时取得最佳 GenImage 验证结果，因此在各基准上统一采用这一学习率。
图~\ref{fig:learning-rate-sensitivity}(b) 显示，在野外基准上，由 GenImage 选出的学习率同样使两种模型在六个候选值中取得最高均值。
}

With the learning rate fixed at $2\times10^{-5}$, we compare random and text initialization of the same linear head (Table~\ref{tab:matched-lr}).
Text initialization improves mean balanced accuracy in the wild by $8.6\%$ for PE and $10.6\%$ for MoEViE.
The gains on the controlled benchmarks are  $2.1\%$ and $3.8\%$, respectively.
\cn{
将学习率固定为 2e-5 后，我们进一步比较同一线性头采用随机初始化和文本初始化时的表现（表~\ref{tab:matched-lr}）。
采用文本初始化后，PE 和 MoEViE 在野外基准上的平均平衡准确率分别提高了 $8.6\%$ 和 $10.6\%$，在控制基准上的增益分别为 $2.1\%$ 和 $3.8\%$。
}

\begin{table}[!t]
\caption{Mean balanced accuracy (\%) for random and text initialization of the linear head at a learning rate of $2\times10^{-5}$. Bold indicates the higher mean for each encoder and benchmark group.}
\cn{学习率为 $2\times10^{-5}$ 时，随机初始化与文本初始化的线性头在两组基准上的平均平衡准确率（\%）。粗体表示各编码器在每组基准上较高的均值。}
\label{tab:matched-lr}
\centering\small
\setlength{\tabcolsep}{3.5pt}
\begin{tabular}{l|l|cc}
\toprule
\textbf{Model} & \textbf{Setting} & \textbf{In-the-Wild} & \textbf{Controlled} \\
\midrule
\multirow{2}{*}{PE} & Linear & 84.3 & 94.4 \\
 & SPC & \textbf{92.9} & \textbf{96.5} \\
\midrule
\multirow{2}{*}{MoEViE} & Linear & 87.9 & 91.7 \\
 & SPC & \textbf{98.5} & \textbf{95.5} \\
\bottomrule
\end{tabular}
\end{table}

\subsection{When Does a VLM Benefit from SPC?}
\cn{5.2 VLM 在什么条件下能够受益于 SPC？}

\begin{figure}[t]
\centering
\includegraphics[width=\columnwidth]{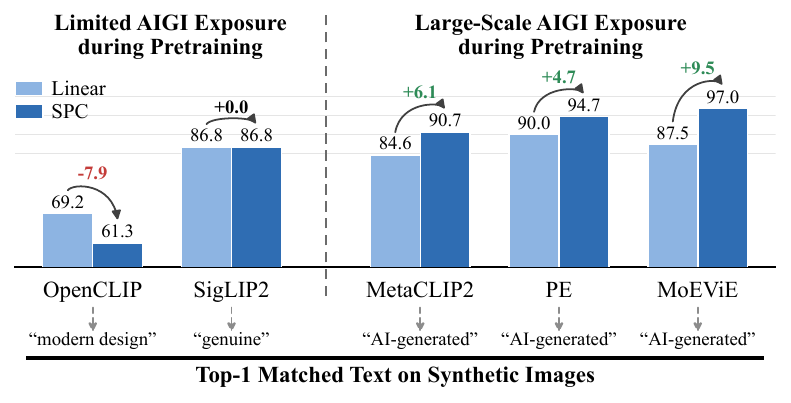}
\caption{Linear probing and SPC on five VLMs, with the same pair of initialization prompts used for SPC throughout. Each score gives equal weight to the in-the-wild and controlled benchmark means.
The label below each model identifies the candidate text concept with the highest average cosine similarity to its frozen image features on generated Chameleon images.
All three models that rank ``AI generated'' first improve under SPC. OpenCLIP declines, while SigLIP2 is unchanged.}
\cn{五种 VLM 在线性探针和 SPC 设置下的表现。所有模型的 SPC 均使用同一对初始化提示词。每个得分由野外基准均值和控制基准均值等权平均得到。模型下方的标签表示与该冻结模型提取的 Chameleon 生成图像特征平均余弦相似度最高的候选文本概念。将“AI generated”排在首位的三个模型均从 SPC 获益，OpenCLIP 的性能下降，SigLIP2 则持平。}
\label{fig:clip-model-comparison}
\end{figure}

We ask why the same initialization benefits some VLMs more than others.
Figure~\ref{fig:clip-model-comparison} compares the change in detection performance with the text concept that each frozen model most closely associates with generated images from Chameleon.
MetaCLIP2, PE, and MoEViE all rank ``AI generated'' first, and all three improve under SPC.
OpenCLIP ranks ``modern design'' first, and its score falls from 69.2\% to 61.3\%. SigLIP2 ranks ``genuine'' first and remains at 86.8\%.
Across these five models, the three that benefit from SPC are precisely those whose pretrained representations already associate generated images with the concept ``AI generated.''
\cn{
我们考察为什么同一种初始化方式对不同 VLM 的效果不同。
图~\ref{fig:clip-model-comparison} 对照了各模型使用 SPC 后的检测性能变化，以及各冻结模型为 Chameleon 生成图像匹配到的首位文本概念。
MetaCLIP2、PE 和 MoEViE 均将“AI generated”排在首位，三者使用 SPC 后性能均有提升。
OpenCLIP 将“modern design”排在首位，得分从 69.2\% 降至 61.3\%。SigLIP2 将“genuine”排在首位，得分保持在 86.8\%。
在这五个模型中，从 SPC 获益的三个模型，恰好都已在预训练表征中建立了生成图像与“AI generated”这一概念之间的联系。
}

\begin{table}[!t]
\caption{Pretraining corpora reported for the evaluated foundation models. Years in parentheses indicate when the dataset or the model report describing the corpus was published.}
\cn{各基础模型报告中列出的预训练语料。括号中的年份表示数据集的发表年份，或介绍该语料的模型报告的发表年份。}
\label{tab:foundation-model-details}
\centering \small
\setlength{\tabcolsep}{3pt}
\begin{tabular}{@{}>{\raggedright\arraybackslash}p{0.30\columnwidth}|>{\raggedright\arraybackslash}p{\dimexpr0.70\columnwidth-2\tabcolsep-\arrayrulewidth\relax}@{}}
\specialrule{\heavyrulewidth}{0pt}{\belowrulesep}
{}\textbf{Model} & \textbf{Reported Pretraining Corpora} \\
\midrule
OpenCLIP & LAION-2B English (2022); LAION-Aesthetics (2022) \\
\midrule
SigLIP2 & WebLI (2023) \\
\midrule
MetaCLIP2 & Worldwide public-web image--text corpus (2025) \\
\midrule
Perception Encoder & Public-web image--text corpus and Video corpus (2025) \\
\midrule
MoEViE & MetaCLIP (2B pairs) and proprietary image--text data (1.5B pairs); additional warmup with CC12M and TreeOfLife (2026 report) \\
\midrule
DINOv3 & LVD-1689M, retrieval-curated web images, and public datasets including ImageNet-1K, ImageNet-22K, and Mapillary Street-level Sequences (2025 report) \\
\bottomrule
\end{tabular}
\end{table}

Differences in pretraining data offer a possible explanation.
As summarized in Table~\ref{tab:foundation-model-details}, OpenCLIP and SigLIP2 were trained on LAION and WebLI, respectively~\cite{OpenCLIP,LAION5B,SigLIP2}. The model reports for MetaCLIP2, PE, and MoEViE describe their use of web image--text corpora~\cite{MetaCLIP2,PE,MoEViE}.
Image--text pairs whose captions identify an image as AI-generated could help a model learn the association observed in Figure~\ref{fig:clip-model-comparison}.
Once that association is present, the embedding of ``AI generated'' gives the classifier an initial direction that already responds to generated images. This may explain why the same prompt improves detection with some encoders but offers little benefit with others.
\cn{
预训练数据的差异为这一现象提供了一种可能的解释。
如表~\ref{tab:foundation-model-details} 所示，OpenCLIP 和 SigLIP2 分别使用 LAION 和 WebLI 训练，MetaCLIP2、PE 和 MoEViE 的模型报告则介绍了各自使用的网络图文语料。
当图文对中的文本明确指出图像由 AI 生成时，这类样本可能帮助模型建立图~\ref{fig:clip-model-comparison} 中观察到的联系。
一旦形成这种联系，“AI generated”的文本嵌入就能为分类器提供一个已经对生成图像有响应的初始方向。这或许可以解释为什么同一句提示词能改善一些编码器的检测性能，却对另一些编码器帮助有限。
}

\begin{table*}[!t]
\caption{Effect of initialization prompts on mean balanced accuracy (\%) for PE and MoEViE. The fake-class and real-class columns list the prompts used to initialize the corresponding prototypes. Each $\Delta$ is computed from the displayed means relative to standard linear probing (No SPC) for the same model and benchmark group. Bold indicates the best result in each column for each model.}
\cn{初始化提示词对 PE 和 MoEViE 平均平衡准确率（\%）的影响。假类和真类两列给出初始化相应类别原型的提示词。每个 Δ 均根据表中均值计算，表示相对于同一模型在同一组基准上采用常规线性探针（No SPC）时的差值。粗体表示每个模型在对应列的最优结果。}
\label{tab:initialization-texts}
\centering \small
\setlength{\tabcolsep}{4pt}
\resizebox{0.8\textwidth}{!}{%
\begin{tabular}{l|l|ll|cc|cc}
\toprule
\textbf{Model} & \textbf{Setting} & \textbf{Fake-class} & \textbf{Real-class} & \textbf{In-the-Wild} & \textbf{$\Delta$} & \textbf{Controlled} & \textbf{$\Delta$} \\
\midrule
\multirow{6}{*}{PE} & No SPC & -- & -- & 84.8 & -- & 95.2 & -- \\
 & Selected Prompt & AI generated & a real photo & \textbf{92.9} & \incre{\textbf{+8.1}} & \textbf{96.5} & \incre{\textbf{+1.3}} \\
 & Similar Prompt & AI art & real & 90.9 & \incre{+6.1} & 95.9 & \incre{+0.7} \\
 & Opposite Prompt & a real photo & AI generated & 38.7 & \decre{-46.1} & 66.3 & \decre{-28.9} \\
 & Unrelated Prompt I & sunset & landscape & 79.1 & \decre{-5.7} & 89.1 & \decre{-6.1} \\
 & Unrelated Prompt II & portrait & technology & 74.8 & \decre{-10.0} & 86.5 & \decre{-8.7} \\
\midrule
\multirow{6}{*}{MoEViE} & No SPC & -- & -- & 81.4 & -- & 93.5 & -- \\
 & Selected Prompt & AI generated & a real photo & \textbf{98.5} & \incre{\textbf{+17.1}} & \textbf{95.5} & \incre{\textbf{+2.0}} \\
 & Similar Prompt & AI art & real & 96.3 & \incre{+14.9} & 95.3 & \incre{+1.8} \\
 & Opposite Prompt & a real photo & AI generated & 37.5 & \decre{-43.9} & 58.6 & \decre{-34.9} \\
 & Unrelated Prompt I & sunset & landscape & 84.3 & \incre{+2.9} & 87.3 & \decre{-6.2} \\
 & Unrelated Prompt II & portrait & technology & 78.4 & \decre{-3.0} & 85.5 & \decre{-8.0} \\
\bottomrule
\end{tabular}
}
\end{table*}

\subsection{How Do Prompt Semantics Drive SPC's Gains?}
\cn{5.1 提示词语义如何带来 SPC 的增益？}

To understand why text initialization improves detection, we vary the prompts used to initialize the real- and fake-class prototypes (Table~\ref{tab:initialization-texts}).
Replacing ``AI generated'' and ``a real photo'' with ``AI art'' and ``real'' preserves most of the gains over linear probing for both encoders on both benchmark groups.
Swapping the original prompts between classes has a much larger effect. With the same two text embeddings, the in-the-wild mean falls from 92.9\% to 38.7\% for PE and from 98.5\% to 37.5\% for MoEViE.
Prompts describing unrelated image content also yield no consistent gain.

This pattern suggests that PE and MoEViE have already learned a visual representation of the concept ``AI generated'' during pretraining, with features of generated images partially aligned with text embeddings that express this concept.
SPC exploits this existing alignment by using descriptions of image provenance to initialize the corresponding class prototypes.
\cn{
为理解文本初始化为何能提升检测性能，我们改变了用于初始化真假类别原型的提示词（表~\ref{tab:initialization-texts}）。
将“AI generated”和“a real photo”换成“AI art”和“real”后，两种编码器在两组基准上都保留了相对于线性探针的大部分增益。
对调原有提示词所对应的类别，影响则大得多。尽管使用的是完全相同的两个文本嵌入，PE 在野外基准上的均值从 92.9\% 降至 38.7\%，MoEViE 也从 98.5\% 降至 37.5\%。
使用与图像来源无关的内容描述，同样未能带来稳定增益。

这些结果提示，PE 和 MoEViE 可能已在预训练中形成了对“AI 生成图像”这一概念的视觉表征，生成图像的特征与表达这一概念的文本嵌入在特征空间中已有一定程度的对齐。
SPC 用描述图像来源的文本嵌入初始化相应的类别原型，使线性分类器从训练之初就能利用这种已有的对齐关系。
}

\begin{figure*}[!t]
\centering
\includegraphics[width=\textwidth]{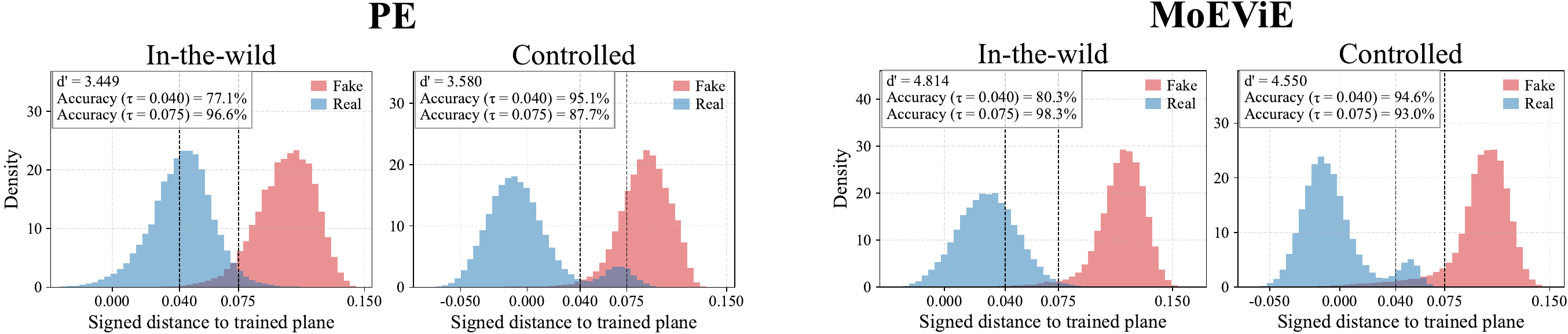}
\caption{Signed distances of real and generated images to the trained single-output plane under fake-only initialization. The left pair of panels shows PE and the right pair shows MoEViE. Within each pair, the in-the-wild distribution is on the left and the controlled distribution is on the right. Vertical dashed lines mark the two decision thresholds, $\tau=0.040$ and $0.075$.}
\cn{仅使用生成类原型初始化时，真实图像与生成图像到训练后单输出超平面的有符号距离分布。左侧两幅为 PE，右侧两幅为 MoEViE。每组内左图为野外基准，右图为控制基准。竖直虚线标出两个决策阈值 τ = 0.040 和 0.075。}
\label{fig:single-output-distance}
\end{figure*}

\subsection{Can a Single-Output Head Replace the Two-Output Head?}
\cn{单输出头能否替代双输出头？}

For binary detection, a two-output head can be expressed as a single-output head because the predicted probability depends only on the difference between the two logits.
Let $n$ be the image feature dimension and $z_c=\mathbf{x}p_c^\top+b_c$ the logit for class $c$. Then
\begin{equation}
\begin{aligned}
P(\mathrm{fake}\mid\mathbf{x})
&=\frac{\exp(z_{\mathrm{fake}})}{\exp(z_{\mathrm{fake}})+\exp(z_{\mathrm{real}})}\\
&=\operatorname{sigmoid}(z_{\mathrm{fake}}-z_{\mathrm{real}})\\
&=\operatorname{sigmoid}(\mathbf{x}w+\beta),\\
w&=p_{\mathrm{fake}}^\top-p_{\mathrm{real}}^\top,\quad
\beta=b_{\mathrm{fake}}-b_{\mathrm{real}}.
\end{aligned}
\label{eq:single-output-equivalence}
\end{equation}
Initializing the $n\times1$ head with these weight and bias differences therefore gives exactly the same initial predictions as the $n\times2$ SPC head.
We call this initialization Diff and compare the two trained heads to determine which generalizes better.
\cn{
对于二分类检测，双输出头可以写成单输出头，因为预测概率仅取决于两个 logit 之差。
设 n 为图像特征维度，类别 c 的 logit 为 z_c = x p_cᵀ + b_c，则

P(fake | x) = exp(z_fake) / [exp(z_fake) + exp(z_real)]
            = sigmoid(z_fake − z_real)
            = sigmoid(xw + β)
w = p_fakeᵀ − p_realᵀ，β = b_fake − b_real

因此，用上述权重差和偏置差初始化 n×1 分类头，其初始预测与 n×2 的 SPC 分类头完全相同。
我们将这种初始化称为 Diff，并比较两种分类头训练后的表现，以判断哪种形式具有更好的泛化性能。
}

\begin{table}[!t]
\caption{Mean balanced accuracy (\%) of the two-output SPC head and two single-output variants. Diff initializes its weight with $p_{\mathrm{fake}}^\top-p_{\mathrm{real}}^\top$ and uses a zero threshold. Fake initializes its weight with $p_{\mathrm{fake}}^\top$ and uses the signed-distance threshold shown in each row. Bold indicates the highest mean for each encoder and benchmark group.}
\cn{双输出 SPC 分类头与两种单输出变体的平均平衡准确率（\%）。Diff 使用 p_fakeᵀ − p_realᵀ 初始化权重，决策阈值为零。Fake 使用 p_fakeᵀ 初始化权重，有符号距离阈值见对应行。粗体表示各编码器在各组基准上的最高均值。}
\label{tab:single-output-ablation}
\centering\small
\setlength{\tabcolsep}{2.5pt}
\resizebox{\columnwidth}{!}{%
\begin{tabular}{l|l|c|cc}
\toprule
\textbf{Model} & \textbf{Setting} & \textbf{Head size} & \textbf{In-the-Wild} & \textbf{Controlled} \\
\midrule
\multirow{4}{*}{PE} & SPC & $n\times2$ & 92.9 & \textbf{96.5} \\
\cmidrule(l){2-5}
 & Diff & \multirow{3}{*}{$n\times1$} & 96.3 & 95.2 \\
 & Fake ($\tau=0.040$) &  & 77.1 & 95.1 \\
 & Fake ($\tau=0.075$) &  & \textbf{96.6} & 87.7 \\
\midrule
\multirow{4}{*}{MoEViE} & SPC & $n\times2$ & 98.5 & \textbf{95.5} \\
\cmidrule(l){2-5}
 & Diff & \multirow{3}{*}{$n\times1$} & \textbf{98.8} & 93.8 \\
 & Fake ($\tau=0.040$) &  & 80.3 & 94.6 \\
 & Fake ($\tau=0.075$) &  & 98.3 & 93.0 \\
\bottomrule
\end{tabular}
}
\end{table}

Table~\ref{tab:single-output-ablation} shows that Diff and SPC achieve similar overall performance after training.
When the in-the-wild and controlled means are weighted equally, Diff performs slightly better for PE, while SPC performs slightly better for MoEViE.
We adopt the two-output formulation because it assigns each class its own prototype and classifies an image by comparing its responses to both.
\cn{
表~\ref{tab:single-output-ablation} 表明，训练后 Diff 与 SPC 的整体性能相近。
对野外和控制两组基准的均值等权平均后，PE 上 Diff 略优，MoEViE 上则是 SPC 略优。
我们采用双输出形式，是因为它更直观。真实类与生成类各有一个原型，检测时直接比较图像与两个原型的匹配程度。
}

Diff still uses both prompts. To examine the role of the real-class reference, we initialize the single-output head with $p_{\mathrm{fake}}^\top$ alone (Fake).
We classify an image as fake when its signed distance to the plane exceeds a threshold $\tau$:
\begin{equation}
d(\mathbf{x})=\frac{\mathbf{x}w+\beta}{\lVert w\rVert_2},\qquad
\widehat{y}=\mathrm{fake}\ \text{if}\ d(\mathbf{x})>\tau.
\label{eq:single-output-threshold}
\end{equation}
With Diff, zero is the natural boundary because the initial score measures the difference between the fake and real responses.
With Fake, many real images also fall on the positive side of the trained plane (Figure~\ref{fig:single-output-distance}).
Moving the threshold to the right helps, but the two benchmark groups favor different thresholds.
For PE, raising $\tau$ from 0.040 to 0.075 improves in-the-wild accuracy from 77.1\% to 96.6\% while reducing controlled accuracy from 95.1\% to 87.7\%. MoEViE shows the same trade-off.
At either tested threshold, Fake falls below SPC when the two groups are weighted equally.
We therefore retain both prompts so that the real-class response provides a reference for the fake-class response without requiring a separately tuned distance threshold.
\cn{
Diff 仍然使用了两条提示词。为考察真实类参照的作用，我们进一步仅用 p_fakeᵀ 初始化单输出头，记为 Fake。
我们计算图像到分类超平面的有符号距离 d(x) = (xw + β) / ‖w‖₂，当 d(x) > τ 时将图像判为生成图像。
对于 Diff，初始分数表示生成类与真实类的响应之差，因此零点是自然的决策边界。
对于 Fake，许多真实图像也落在训练后超平面的正侧（图~\ref{fig:single-output-distance}）。
将阈值右移有助于区分两类图像，但两组基准偏好的阈值不同。
以 PE 为例，将 τ 从 0.040 提高到 0.075 后，野外准确率从 77.1\% 升至 96.6\%，控制基准准确率却从 95.1\% 降至 87.7\%。MoEViE 也呈现相同趋势。
对两组基准等权平均后，Fake 在两个测试阈值下的整体表现均低于 SPC。
因此，我们保留两条提示词，以真实类响应作为判断生成类响应的参照，省去单独调节距离阈值的步骤。
}

\subsection{Can Simpler Alternatives Replace SPC?}
\cn{更朴素的替代方法能否取代 SPC？}

SPC achieves its gains through a subtle change to linear probing. We explore whether a more straightforward alternative could achieve similar gains, taking a larger classifier head as an example.
We replace the randomly initialized linear head with a randomly initialized MLP of the form
\begin{equation}
\operatorname{Linear}(n,d_h)
\;\rightarrow\;\operatorname{ReLU}
\;\rightarrow\;\operatorname{Linear}(d_h,2).
\label{eq:mlp-head}
\end{equation}

Here, $n$ is the image feature dimension and $d_h$ is the hidden dimension, which we set to 512. The two outputs are the real- and fake-class logits.
\cn{
SPC 通过对线性探针的一处巧妙改动提升了检测性能。我们进一步考察是否有更朴素的做法也能取得类似收益，例如单纯增大分类头。为此，我们将随机初始化的线性头替换为同样随机初始化的 MLP，其结构为

Linear(n, d_h) → ReLU → Linear(d_h, 2)

其中，n 为图像特征维度，d_h 为隐藏层维度，在实验中设为 512。两个输出分别为真实类和生成类的 logit。
}

\begin{table}[ht]
\caption{Mean balanced accuracy (\%) on the in-the-wild and controlled benchmarks for three classifier heads trained on frozen PE and MoEViE features. Linear and MLP use random initialization. SPC uses a linear head initialized with text prototypes. Bold indicates the highest mean for each encoder and benchmark group.}
\cn{三种分类头在野外和控制基准上的平均平衡准确率（\%），均使用冻结的 PE 或 MoEViE 图像特征训练。Linear 和 MLP 采用随机初始化，SPC 使用由文本原型初始化的线性头。粗体表示各编码器在各组基准上的最高均值。}
\label{tab:mlp-ablation}
\centering\small
\setlength{\tabcolsep}{3.5pt}
\begin{tabular}{l|l|cc}
\toprule
\textbf{Model} & \textbf{Setting} & \textbf{In-the-Wild} & \textbf{Controlled} \\
\midrule
\multirow{3}{*}{PE} & Linear & 84.8 & 95.2 \\
 & MLP & 85.0 & 90.1 \\
 & SPC & \textbf{92.9} & \textbf{96.5} \\
\midrule
\multirow{3}{*}{MoEViE} & Linear & 81.4 & 93.5 \\
 & MLP & 90.2 & \textbf{95.5} \\
 & SPC & \textbf{98.5} & \textbf{95.5} \\
\bottomrule
\end{tabular}
\end{table}

Table~\ref{tab:mlp-ablation} shows that SPC achieves better in-the-wild detection than the larger MLP head with both frozen encoders.
On PE, SPC averages 92.9\% in the wild, compared with 85.0\% for the MLP. On MoEViE, the corresponding scores are 98.5\% and 90.2\%.
On the controlled benchmarks, SPC also outperforms the MLP on PE (96.5\% vs.\ 90.1\%) and matches it on MoEViE (95.5\%).
These results suggest that the frozen features already support a linear decision boundary that generalizes well.
For both encoders, initializing the linear head with real and fake text prototypes yields better transfer to in-the-wild images than increasing classifier capacity alone.
\cn{
表~\ref{tab:mlp-ablation} 显示，在两种冻结编码器上，SPC 的野外检测性能均优于容量更大的 MLP 分类头。
在 PE 上，SPC 的野外均值为 92.9\%，高于 MLP 的 85.0\%。在 MoEViE 上，两者分别为 98.5\% 和 90.2\%。
在控制基准上，SPC 在 PE 上同样优于 MLP（96.5\% 对 90.1\%），在 MoEViE 上则与其持平（95.5\%）。
这些结果表明，冻结的图像特征已经支持具有良好泛化能力的线性决策边界。
对于两种编码器，使用真假文本原型初始化线性头，都比单纯增加分类器容量更有利于泛化到野外图像。
}

\begin{figure}[t]
\centering
\includegraphics[width=\columnwidth]{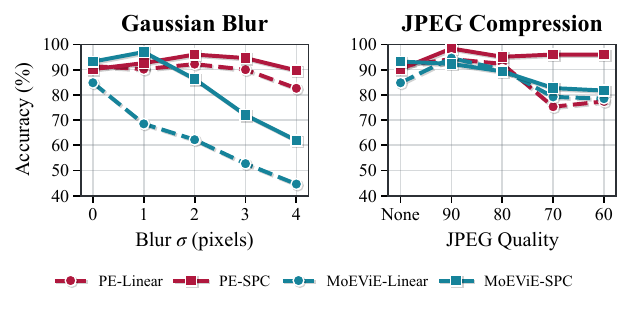}
\caption{Balanced accuracy (\%) of PE and MoEViE on UnivFakeDetect under Gaussian blur (left) and JPEG compression (right). Dashed lines with circles denote standard linear probing, and solid lines with squares denote SPC. A blur strength of zero and ``None'' denote unperturbed images.}
\cn{PE 和 MoEViE 在 UnivFakeDetect 上经过高斯模糊（左）和 JPEG 压缩（右）后的平衡准确率（\%）。带圆点的虚线表示常规线性探针，带方点的实线表示 SPC。左图的模糊强度 0 和右图的“None”均表示未经扰动的图像。}
\label{fig:common-perturbations}
\end{figure}

\subsection{Resilience to Common Perturbations}
\cn{5.4 对常见扰动的鲁棒性}

Figure~\ref{fig:common-perturbations} compares SPC with standard linear probing on UnivFakeDetect under Gaussian blur and JPEG compression.
Averaged over the eight perturbed conditions, SPC raises the score from 86.8\% to 94.8\% for PE and from 71.3\% to 82.9\% for MoEViE.
Both encoders thus retain an overall advantage over their linear baselines after image post-processing.
\cn{
图~\ref{fig:common-perturbations} 比较了 SPC 与常规线性探针在 UnivFakeDetect 上应对高斯模糊和 JPEG 压缩时的表现。
在八种扰动条件下取平均，SPC 将 PE 的得分从 86.8\% 提升至 94.8\%，将 MoEViE 的得分从 71.3\% 提升至 82.9\%。
这表明，图像经过后处理后，采用 SPC 的两种编码器在整体上仍优于各自的线性探针基线。
}

\section{Conclusion}
\cn{6. 结论}

PE's frozen features exhibit a clearer forensic structure than those of DINOv3, yet PE performs worse under conventional linear probing.
SPC makes better use of this structure by initializing the linear head of the frozen encoder with text embeddings that describe real and generated images.
This change substantially improves in-the-wild detection for both PE and MoEViE and also improves their overall performance on the controlled benchmarks.
Across nine in-the-wild subsets, MoEViE-SPC averages 98.5\%, surpassing all specialized detectors in our comparison, including those with more elaborate task-specific designs.
These results establish SPC as a simple and effective baseline for future AIGI detection research.
\cn{
PE 的冻结特征比 DINOv3 的冻结特征呈现出更清晰的取证结构，但采用常规线性探针时，PE 的表现反而更差。
SPC 利用描述真实图像和生成图像的文本嵌入初始化冻结编码器的线性头，使分类器能够更充分地利用这一结构。
这一改动大幅提升了 PE 和 MoEViE 的野外检测性能，也改善了两者在控制基准上的整体表现。
在九个野外子集上，MoEViE-SPC 的均值达到 98.5\%，超过了参与比较的所有专用检测器，包括采用更复杂任务专用设计的方法。
这些结果为后续 AIGI 检测研究确立了一个简单有效的基线。
}

\FloatBarrier
\bibliographystyle{IEEEtran}
\bibliography{references}
\FloatBarrier

\clearpage
\appendices

\section{Hardware and Software Environment}
\cn{硬件与软件环境}

All experiments were run on a system equipped with an Intel Xeon Gold 6226R CPU and an NVIDIA Tesla V100S GPU with 32~GB of GPU memory.
We used PyTorch 2.2.0 with CUDA 11.8, cuDNN 8.7, and NVIDIA driver version 510.47.03.
\cn{
所有实验均在配备 Intel Xeon Gold 6226R CPU 和 NVIDIA Tesla V100S GPU（32 GB 显存）的系统上运行。
软件环境为 PyTorch 2.2.0、CUDA 11.8、cuDNN 8.7 和 NVIDIA 510.47.03 驱动。
}

\section{Text-Concept Matching Protocol}
\cn{文本概念匹配协议}

To investigate why some VLMs benefit more from SPC than others, we test whether their frozen representations already associate the visual features of generated images with the concept ``AI generated.''
Following the image--text matching protocol of \emph{Simplicity Prevails}~\cite{SimplicityPrevails}, we compare candidates from three groups.
\begin{itemize}
	\item \textbf{Forensic concepts:} terms describing whether an image is real, manipulated, or AI-generated, including ``fake,'' ``real,'' ``AI generated,'' ``authentic,'' ``genuine,'' ``manipulated,'' and ``synthetic.''
	\item \textbf{Content concepts:} terms describing what an image depicts without referring to how it was produced, including ``sunset,'' ``landscape,'' ``portrait,'' ``abstract art,'' ``modern design,'' ``technology,'' and ``nature.''
	\item \textbf{Source concepts:} names of generative models, datasets, or platforms, including ``GenImage,'' ``ADM,'' ``BigGAN,'' ``glide,'' and ``Midjourney.''
\end{itemize}
\cn{
为探究为什么部分 VLM 能从 SPC 中获得更大收益，我们检验其冻结表征是否已经将生成图像的视觉特征与“AI generated”这一概念联系起来。
我们沿用 Simplicity Prevails~\cite{SimplicityPrevails} 的图文匹配协议，比较以下三组候选文本。

- 伪造相关概念：明确描述真实性或人工生成属性的词语，包括“fake”“real”“AI generated”“authentic”“genuine”“manipulated”和“synthetic”。
- 内容相关概念：对视觉内容的中性描述，包括“sunset”“landscape”“portrait”“abstract art”“modern design”“technology”和“nature”。
- 来源相关概念：生成模型、数据集或平台的名称，包括“GenImage”“ADM”“BigGAN”“glide”和“Midjourney”。
}

\begin{figure}[t]
\centering
\includegraphics[width=\columnwidth]{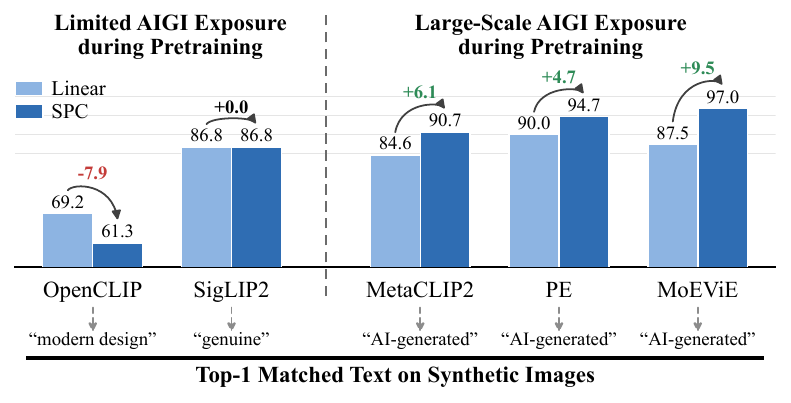}
\caption{Linear probing and SPC on five VLMs, with the same pair of initialization prompts used for SPC throughout. Each score gives equal weight to the in-the-wild and controlled benchmark means.
The label below each model identifies the candidate text concept with the highest average cosine similarity to its frozen image features on generated Chameleon images.
All three models that rank ``AI generated'' first improve under SPC. OpenCLIP declines, while SigLIP2 is unchanged.}
\cn{五种 VLM 在线性探针和 SPC 设置下的表现。所有模型的 SPC 均使用同一对初始化提示词。每个得分由野外基准均值和控制基准均值等权平均得到。模型下方的标签表示与该冻结模型提取的 Chameleon 生成图像特征平均余弦相似度最高的候选文本概念。将“AI generated”排在首位的三个模型均从 SPC 获益，OpenCLIP 的性能下降，SigLIP2 则持平。}
\label{fig:appendix-clip-model-comparison}
\end{figure}

For each frozen model, we compute the cosine similarity between each candidate text embedding and the image embedding of every generated image in Chameleon.
Figure~\ref{fig:appendix-clip-model-comparison} reports the concept with the highest mean similarity across images.
MetaCLIP2, PE, and MoEViE all rank ``AI generated'' first, whereas OpenCLIP ranks ``modern design'' first and SigLIP2 ranks ``genuine'' first.
For the first three models, the visual features of generated images thus align most closely with the candidate text that correctly describes their origin.
This pattern is consistent with the hypothesis that repeated exposure during pretraining to AI-generated images paired with captions identifying them as such helps models learn this association.
Such alignment may explain why ``AI generated'' provides a useful starting point for SPC on these models: its text embedding responds to generated images before calibration.
\cn{
对于每个冻结模型，我们计算各候选文本嵌入与 Chameleon 中每张生成图像的图像嵌入之间的余弦相似度，再对图像取平均。
图~\ref{fig:appendix-clip-model-comparison} 给出了各模型平均相似度最高的概念。
MetaCLIP2、PE 和 MoEViE 均将“AI generated”排在首位，而 OpenCLIP 和 SigLIP2 的首位匹配分别是“modern design”和“genuine”。
对于前三个模型，生成图像的视觉特征与正确描述其来源的候选文本最为接近。
这一现象与第 5.2 节提出的解释一致：如果模型在预训练阶段反复接触生成图像及明确指出其 AI 来源的配对文本，就可能学会将这类视觉特征与“AI generated”联系起来。
这种对齐也有助于解释为什么该提示词能为这三个模型的 SPC 提供有效起点：在原型校准之前，它的文本嵌入就能够对生成图像产生响应。
}

\begin{table*}[!t]
\caption{Balanced accuracy (\%) on GenImage~\cite{GenImage}. Avg. is the unweighted mean across the eight generator subsets, including SD-v1.4. Bold indicates the highest value in each column.}
\cn{GenImage~\cite{GenImage} 上的平衡准确率（\%）。Avg. 为包含 SD-v1.4 在内的八个生成器子集的等权平均值。粗体表示各列最高值。}
\label{tab:genimage}
\centering
\setlength{\tabcolsep}{2mm}
\resizebox{\textwidth}{!}{%
\begin{tabular}{l|cccccccc|c}
\toprule
\textbf{Method} & \textbf{ADM} & \textbf{BigGAN} & \textbf{Midjourney} & \textbf{VQDM} & \textbf{GLIDE} & \textbf{SD-v1.4} & \textbf{SD-v1.5} & \textbf{Wukong} & \textbf{Avg.} \\
\midrule
\multicolumn{10}{c}{\textit{Traditional AIGI Detectors}} \\
\midrule
NPR~\cite{NPR} & 43.8 & 57.7 & 53.4 & 48.4 & 41.2 & 55.1 & 55.0 & 57.4 & 51.5 \\
AIDE~\cite{AIDE} & 50.4 & 50.6 & 58.2 & 50.8 & 54.6 & 77.2 & 77.4 & 70.5 & 61.2 \\
DDA~\cite{DDA} & 89.5 & 86.5 & 95.6 & 76.5 & 89.6 & 98.7 & 98.6 & 98.7 & 91.7 \\
\midrule
\multicolumn{10}{c}{\textit{Specialized AIGI Detectors with Modern VFMs}} \\
\midrule
FGTS~\cite{FGTS} & 90.7 & 94.4 & 85.7 & 93.5 & 92.0 & 98.1 & 92.9 & 93.6 & 92.6 \\
\midrule
\multicolumn{10}{c}{\textit{Frozen VFM + Linear Probing}} \\
\midrule
OpenCLIP-Linear & 55.1 & 50.6 & 87.5 & 62.1 & 70.7 & 98.5 & 98.4 & 91.8 & 76.8 \\
SigLIP2-Linear & 86.1 & 90.4 & 86.1 & 94.4 & 95.7 & 99.8 & 99.6 & 99.4 & 93.9 \\
MetaCLIP2-Linear & 74.1 & 79.6 & 97.8 & 85.8 & 89.6 & 99.7 & 99.6 & 98.3 & 90.6 \\
PE-Linear & 85.9 & 98.4 & 96.8 & 99.2 & 97.7 & \textbf{100.0} & \textbf{99.9} & \textbf{100.0} & 97.2 \\
DINOv3-Linear & 85.5 & \textbf{99.3} & 93.6 & 99.2 & 96.8 & 99.9 & 99.8 & 99.6 & 96.7 \\
MoEViE-Linear & 81.5 & 91.4 & \textbf{98.8} & 97.8 & 94.2 & \textbf{100.0} & \textbf{99.9} & \textbf{100.0} & 95.4 \\
\midrule
\multicolumn{10}{c}{\textit{Frozen VFM + SPC (Ours)}} \\
\midrule
PE-SPC (Ours) & \textbf{93.6} & \textbf{99.3} & 97.7 & \textbf{99.6} & \textbf{98.9} & 99.8 & 99.8 & 99.8 & \textbf{98.6} \\
MoEViE-SPC (Ours) & 83.4 & 94.9 & 98.0 & 97.9 & 94.8 & 99.9 & \textbf{99.9} & \textbf{100.0} & 96.1 \\
\bottomrule
\end{tabular}
}
\end{table*}

\begin{table*}[!t]
\caption{Balanced accuracy (\%) on UnivFakeDetect~\cite{UnivFD} under the evaluation protocol of MIRROR~\cite{MIRROR}. Avg. is the unweighted mean across the evaluated subsets. G denotes GLIDE, and LDM-200-C denotes \texttt{ldm\_200\_cfg}. Bold indicates the highest value in each column.}
\cn{沿用 MIRROR~\cite{MIRROR} 的评测设置，在 UnivFakeDetect~\cite{UnivFD} 上得到的平衡准确率（\%）。Avg. 为各评测子集的等权平均值。G 表示 GLIDE，LDM-200-C 表示 ldm\_200\_cfg。粗体表示各列最高值。}
\label{tab:univfakedetect}
\centering \small
\setlength{\tabcolsep}{1.2mm}
\resizebox{\textwidth}{!}{%
\begin{tabular}{l|*{12}{c}|c}
\toprule
\textbf{Method} & \textbf{SITD} & \textbf{SAN} & \textbf{CRN} & \textbf{IMLE} & \textbf{Guided} & \textbf{LDM-200} & \textbf{LDM-200-C} & \textbf{LDM-100} & \textbf{G-100/27} & \textbf{G-50/27} & \textbf{G-100/10} & \textbf{DALL-E} & \textbf{Avg.} \\
\midrule
\multicolumn{14}{c}{\textit{Traditional AIGI Detectors}} \\
\midrule
NPR~\cite{NPR} & 57.8 & 36.8 & 52.7 & 52.6 & 78.5 & 50.4 & 50.4 & 49.6 & 46.9 & 50.9 & 44.2 & 43.3 & 51.1 \\
AIDE~\cite{AIDE} & 55.0 & 60.5 & 53.1 & 76.4 & 87.4 & 93.5 & 85.6 & 93.7 & 78.3 & 78.1 & 81.1 & 91.7 & 77.9 \\
DDA~\cite{DDA} & 48.9 & \textbf{87.0} & 60.8 & 67.5 & 88.1 & 73.7 & 73.5 & 72.9 & 69.1 & 67.9 & 71.6 & 51.7 & 69.4 \\
\midrule
\multicolumn{14}{c}{\textit{Specialized AIGI Detectors with Modern VFMs}} \\
\midrule
FGTS~\cite{FGTS} & 63.3 & 61.4 & 96.8 & 98.0 & 91.0 & 94.9 & 94.7 & 94.9 & 94.6 & 94.6 & 94.2 & 94.8 & 89.4 \\
MIRROR~\cite{MIRROR} & 63.3 & 80.6 & 76.4 & 76.9 & 79.8 & 98.9 & 98.8 & 99.0 & 95.1 & 94.7 & 95.6 & 98.6 & 88.2 \\
\midrule
\multicolumn{14}{c}{\textit{Frozen VFM + Linear Probing}} \\
\midrule
OpenCLIP-Linear & 64.4 & 68.3 & 39.2 & 32.6 & 64.1 & 74.3 & 75.7 & 75.4 & 66.6 & 65.9 & 68.4 & 69.8 & 63.7 \\
SigLIP2-Linear & 66.1 & 67.6 & 67.6 & 72.5 & 89.1 & 97.3 & 96.7 & 97.3 & 95.9 & 95.4 & 95.3 & 96.4 & 86.4 \\
MetaCLIP2-Linear & 73.9 & 84.9 & 53.1 & 53.3 & 76.9 & 97.5 & 94.2 & 97.7 & 94.4 & 95.3 & 95.5 & 94.9 & 84.3 \\
PE-Linear & \textbf{88.9} & 64.2 & 78.5 & 78.5 & 91.8 & \textbf{100.0} & \textbf{99.8} & \textbf{100.0} & 98.7 & 98.5 & 98.2 & \textbf{100.0} & 91.4 \\
DINOv3-Linear & 60.0 & 61.6 & 93.9 & 95.3 & 88.9 & 99.5 & 98.7 & 99.4 & 98.8 & 98.8 & 98.9 & 99.0 & 91.1 \\
MoEViE-Linear & 66.7 & 76.5 & 51.8 & 51.8 & 80.7 & 99.4 & 99.4 & 99.5 & 97.3 & 97.4 & 97.8 & 99.4 & 84.8 \\
\midrule
\multicolumn{14}{c}{\textit{Frozen VFM + SPC (Ours)}} \\
\midrule
PE-SPC (Ours) & 76.1 & 73.1 & 70.3 & 70.3 & \textbf{95.4} & 99.7 & 99.6 & 99.7 & \textbf{99.6} & \textbf{99.5} & \textbf{99.5} & 99.7 & 90.2 \\
MoEViE-SPC (Ours) & 78.1 & 70.3 & \textbf{98.4} & \textbf{98.4} & 81.7 & 99.9 & 99.4 & 99.9 & 98.0 & 98.0 & 98.3 & 98.7 & \textbf{93.2} \\
\bottomrule
\end{tabular}
}
\end{table*}

\begin{table*}[!t]
\caption{Balanced accuracy (\%) on AIGI-Now~\cite{AIGINow}. For each generator, pix uses format-aligned images, while sem applies degradations to suppress low-level generation traces. Avg. is the unweighted mean across both splits for all nine generators. Kontext denotes FLUX-Kontext. Bold indicates the highest value in each column.}
\cn{AIGI-Now~\cite{AIGINow} 上的平衡准确率（\%）。对于每个生成器，pix 使用格式对齐的图像，sem 通过退化处理削弱低层生成痕迹。Avg. 为九个生成器在两种划分上共十八项结果的等权平均值。Kontext 表示 FLUX-Kontext。粗体表示各列最高值。}
\label{tab:aigi-now}
\centering \small
\setlength{\tabcolsep}{0.6mm}
\resizebox{\textwidth}{!}{%
\begin{tabular}{l|*{9}{cc|}c}
\toprule
\multirow{2}{*}{\textbf{Detector}} &
\multicolumn{2}{c|}{\textbf{FLUX-dev}} &
\multicolumn{2}{c|}{\textbf{FLUX-kera}} &
\multicolumn{2}{c|}{\textbf{Kontext}} &
\multicolumn{2}{c|}{\textbf{FLUX-pro}} &
\multicolumn{2}{c|}{\textbf{gpt4o}} &
\multicolumn{2}{c|}{\textbf{jimeng}} &
\multicolumn{2}{c|}{\textbf{keling}} &
\multicolumn{2}{c|}{\textbf{minimax}} &
\multicolumn{2}{c|}{\textbf{Nano}} &
\multirow{2}{*}{\textbf{Avg.}} \\
\cmidrule(lr){2-19}
& pix & sem & pix & sem & pix & sem & pix & sem & pix & sem & pix & sem & pix & sem & pix & sem & pix & sem & \\
\midrule
\multicolumn{20}{c}{\textit{Traditional AIGI Detectors}} \\
\midrule
NPR~\cite{NPR} & 94.4 & 50.0 & 50.8 & 50.0 & 78.5 & 50.2 & 50.2 & 50.2 & 96.6 & 50.0 & 49.7 & 50.1 & 95.7 & 50.0 & 54.8 & 50.0 & 93.0 & 50.0 & 61.9 \\
AIDE~\cite{AIDE} & 99.1 & 59.0 & 50.4 & 56.9 & \textbf{97.9} & 80.6 & 60.1 & 53.8 & 74.7 & 51.8 & 63.9 & 51.4 & 98.2 & 55.4 & 51.4 & 54.1 & 98.9 & 51.8 & 67.2 \\
DDA~\cite{DDA} & 91.6 & 51.2 & 59.4 & 49.9 & 82.7 & 52.9 & 76.6 & 55.0 & 92.3 & 65.4 & 87.0 & 65.4 & 96.1 & 64.6 & 83.3 & 50.5 & 81.6 & 56.2 & 70.1 \\
\midrule
\multicolumn{20}{c}{\textit{Specialized AIGI Detectors with Modern VFMs}} \\
\midrule
FGTS~\cite{FGTS} & 93.5 & 91.4 & 84.1 & 69.0 & 75.5 & 70.7 & 77.3 & 90.5 & 91.4 & 92.1 & 81.3 & 90.1 & 88.2 & 87.4 & 73.0 & 71.2 & 90.3 & 84.8 & 83.4 \\
ARA~\cite{ARA} & 91.3 & 95.8 & 78.0 & 84.5 & 77.1 & 84.9 & 84.6 & 94.7 & 83.1 & 97.0 & 85.4 & 95.8 & 88.8 & 94.6 & 68.4 & 82.7 & 87.0 & 94.5 & 87.1 \\
\midrule
\multicolumn{20}{c}{\textit{Frozen VFM + Linear Probing}} \\
\midrule
OpenCLIP-Linear & 97.0 & 84.8 & 97.5 & 85.3 & 73.6 & 69.0 & 89.4 & 81.2 & 95.9 & 78.5 & 82.9 & 76.0 & 97.5 & 88.7 & 95.4 & 78.6 & 97.5 & 84.3 & 86.3 \\
SigLIP2-Linear & 95.4 & 86.8 & 90.0 & 69.9 & 79.5 & 69.0 & 87.9 & 88.4 & 94.3 & 75.9 & 81.8 & 81.4 & 95.1 & 86.0 & 85.6 & 68.3 & 90.7 & 87.2 & 84.1 \\
MetaCLIP2-Linear & \textbf{99.7} & 88.0 & 98.4 & 79.5 & 85.5 & 76.6 & \textbf{98.6} & 92.7 & 98.7 & 88.6 & 99.2 & 89.4 & \textbf{99.1} & 89.3 & 96.9 & 77.3 & 99.6 & 88.5 & 91.4 \\
PE-Linear & \textbf{99.7} & 93.6 & 98.9 & 89.8 & 93.3 & 88.3 & 97.8 & 94.3 & 98.8 & 94.8 & 97.6 & 90.1 & 97.6 & 93.1 & \textbf{98.0} & 87.5 & \textbf{99.9} & 90.4 & 94.6 \\
DINOv3-Linear & 94.2 & 94.9 & 86.3 & 77.3 & 73.2 & 73.2 & 81.2 & 93.5 & 90.2 & 94.2 & 82.1 & 92.1 & 89.7 & 90.4 & 74.3 & 76.0 & 90.2 & 88.8 & 85.6 \\
MoEViE-Linear & 99.5 & 98.2 & \textbf{99.7} & 82.1 & 90.8 & 78.7 & 96.0 & 97.5 & \textbf{99.3} & 97.2 & \textbf{99.4} & 98.1 & 98.9 & 91.9 & 95.8 & 80.6 & \textbf{99.9} & 92.6 & 94.2 \\
\midrule
\multicolumn{20}{c}{\textit{Frozen VFM + SPC (Ours)}} \\
\midrule
PE-SPC (Ours) & 99.3 & \textbf{99.8} & 99.0 & \textbf{97.8} & 94.0 & \textbf{96.3} & 97.2 & \textbf{98.6} & 99.1 & \textbf{99.7} & 96.9 & \textbf{99.5} & 98.3 & \textbf{97.6} & 97.4 & \textbf{96.1} & 99.1 & \textbf{99.1} & \textbf{98.0} \\
MoEViE-SPC (Ours) & 98.7 & 98.4 & 97.6 & 80.2 & 86.6 & 79.3 & 93.7 & 96.8 & 97.9 & 98.0 & 99.1 & 98.1 & 97.2 & 92.1 & 89.8 & 78.9 & 99.3 & 94.2 & 93.1 \\
\bottomrule
\end{tabular}
}
\end{table*}

\begin{table*}[!t]
\caption{Balanced accuracy (\%) on AIGI-Holmes~\cite{AIGIHolmes}. Avg. is the mean across the ten generator subsets. JP-1B, JP-7B, LGen, and PA-XL denote Janus-Pro-1B, Janus-Pro-7B, LlamaGen, and PixArt-XL, respectively. Bold indicates the highest value in each column.}
\cn{AIGI-Holmes~\cite{AIGIHolmes} 上的平衡准确率（\%）。Avg. 为十个生成器子集的等权平均值。JP-1B、JP-7B、LGen 和 PA-XL 分别表示 Janus-Pro-1B、Janus-Pro-7B、LlamaGen 和 PixArt-XL。粗体表示各列最高值。}
\label{tab:aigi-holmes}
\centering \small
\setlength{\tabcolsep}{2mm}
\resizebox{\textwidth}{!}{%
\begin{tabular}{l|*{10}{c}|c}
\toprule
\textbf{Detector} & \textbf{FLUX} & \textbf{Infinity} & \textbf{JP-1B} & \textbf{JP-7B} & \textbf{Janus} & \textbf{LGen} & \textbf{PA-XL} & \textbf{SD3.5-L} & \textbf{Show-o} & \textbf{VAR} & \textbf{Avg.} \\
\midrule
\multicolumn{12}{c}{\textit{Traditional AIGI Detectors}} \\
\midrule
NPR~\cite{NPR} & 96.8 & 98.3 & 49.5 & 49.5 & 49.5 & 94.7 & 89.9 & 93.7 & 98.8 & 73.0 & 79.4 \\
AIDE~\cite{AIDE} & 94.4 & 98.7 & 98.9 & 97.8 & 91.2 & 99.4 & 98.6 & 99.4 & 98.0 & 93.6 & 97.0 \\
DDA~\cite{DDA} & 97.2 & 98.9 & 99.3 & 98.7 & 98.5 & 99.3 & 99.4 & 97.0 & 94.8 & 80.4 & 96.4 \\
\midrule
\multicolumn{12}{c}{\textit{Specialized AIGI Detectors with Modern VFMs}} \\
\midrule
FGTS~\cite{FGTS} & 89.9 & 97.8 & 97.1 & 96.3 & 97.2 & 97.8 & 97.7 & 92.1 & 97.7 & 96.0 & 96.0 \\
ARA~\cite{ARA} & 91.0 & 99.7 & 99.6 & 99.8 & 99.8 & 99.7 & 99.8 & 93.5 & 99.6 & \textbf{99.7} & 98.2 \\
\midrule
\multicolumn{12}{c}{\textit{Frozen VFM + Linear Probing}} \\
\midrule
OpenCLIP-Linear & 97.1 & 97.6 & 65.0 & 57.9 & 48.6 & 97.4 & 96.2 & 95.3 & 97.0 & 65.5 & 81.8 \\
SigLIP2-Linear & 95.8 & 99.7 & 98.8 & 97.5 & 97.9 & 99.7 & 99.8 & 91.6 & 99.6 & 89.1 & 96.9 \\
MetaCLIP2-Linear & 99.8 & 99.8 & 93.0 & 88.3 & 80.9 & 99.7 & 99.8 & 99.4 & 99.8 & 73.3 & 93.4 \\
PE-Linear & 99.7 & \textbf{100.0} & 95.1 & 93.1 & 91.7 & 99.9 & \textbf{100.0} & 98.6 & 99.9 & 98.4 & 97.6 \\
DINOv3-Linear & 94.3 & 99.9 & 99.3 & 98.6 & 99.6 & 99.9 & 99.9 & 90.5 & 99.7 & 92.7 & 97.4 \\
MoEViE-Linear & \textbf{99.9} & \textbf{100.0} & 99.7 & 99.9 & 99.7 & \textbf{100.0} & \textbf{100.0} & \textbf{99.9} & \textbf{100.0} & 94.6 & 99.4 \\
\midrule
\multicolumn{12}{c}{\textit{Frozen VFM + SPC (Ours)}} \\
\midrule
PE-SPC (Ours) & 99.3 & 99.5 & 99.3 & 98.8 & 98.5 & 99.5 & 99.4 & 99.1 & 99.4 & 98.9 & 99.2 \\
MoEViE-SPC (Ours) & 99.4 & \textbf{100.0} & \textbf{99.8} & \textbf{100.0} & \textbf{99.9} & \textbf{100.0} & \textbf{100.0} & 99.4 & \textbf{100.0} & 96.6 & \textbf{99.5} \\
\bottomrule
\end{tabular}
}
\end{table*}

\section{Performance on Controlled Benchmarks}
\cn{控制基准上的性能}

\subsection{Generalization on GenImage}
\cn{GenImage 上的泛化性能}

GenImage~\cite{GenImage} allows us to compare performance on the training generator, SD-v1.4, with transfer to seven unseen generators.
As shown in Table~\ref{tab:genimage}, PE already achieves near-perfect scores on SD-v1.4 with either head.
Its largest gain from SPC comes on ADM, its weakest subset under linear probing, where its score rises from 85.9\% to 93.6\%.
Across all eight subsets, SPC raises PE's mean from 97.2\% to 98.6\%, the highest among the compared methods.
MoEViE also improves, from 95.4\% to 96.1\% on average, with its largest gain on BigGAN (91.4\% to 94.9\%).
\cn{
GenImage~\cite{GenImage} 使我们能够比较检测器在训练生成器 SD-v1.4 上的表现，以及它向七个未见生成器的泛化能力。
如表~\ref{tab:genimage} 所示，PE 采用两种分类头时，在 SD-v1.4 上都已接近满分。
SPC 带来的最大提升出现在 PE 线性探针表现最弱的 ADM 上，得分从 85.9\% 升至 93.6\%。
在全部八个子集上，SPC 将 PE 的均值从 97.2\% 提升至 98.6\%，为参与比较的方法中最高。
MoEViE 的均值也从 95.4\% 提升至 96.1\%，其中提升最大的是 BigGAN，从 91.4\% 升至 94.9\%。
}

\subsection{Generalization on UnivFakeDetect}
\cn{UnivFakeDetect 上的泛化性能}

We evaluate on UnivFakeDetect~\cite{UnivFD} following the protocol used by MIRROR~\cite{MIRROR}.
Table~\ref{tab:univfakedetect} shows that SPC raises MoEViE's mean from 84.8\% to 93.2\%, the highest among the compared methods.
CRN and IMLE account for most of this improvement. On each subset, MoEViE-Linear performs near chance at 51.8\%, while MoEViE-SPC reaches 98.4\% using the same frozen features.
\cn{
我们沿用 MIRROR~\cite{MIRROR} 的评测设置，在 UnivFakeDetect~\cite{UnivFD} 上进行比较。
如表~\ref{tab:univfakedetect} 所示，SPC 将 MoEViE 的均值从 84.8\% 提升至 93.2\%，为参与比较的方法中最高。
这一提升主要来自 CRN 和 IMLE。在这两个子集上，MoEViE-Linear 的得分均为 51.8\%，接近随机水平，而使用相同冻结特征的 MoEViE-SPC 均达到 98.4\%。
}

\subsection{Generalization to Recent Generators}
\cn{面向近期生成器的泛化性能}

We evaluate whether a detector trained on SD-v1.4 generalizes to recent generators using AIGI-Now~\cite{AIGINow} and AIGI-Holmes~\cite{AIGIHolmes}.
On AIGI-Now, PE-SPC ranks first with a mean of 98.0\%, compared with 94.6\% for PE-Linear and 85.6\% for DINOv3-Linear (Table~\ref{tab:aigi-now}).
The gain comes from the sem split, where degradation suppresses low-level generation traces.
SPC improves PE on all nine sem subsets, raising their mean from 91.3\% to 98.3\%, while the pix mean changes little (98.0\% to 97.8\%).
This suggests that SPC helps PE distinguish generated images when low-level cues are weakened.
\cn{
我们在 AIGI-Now~\cite{AIGINow} 和 AIGI-Holmes~\cite{AIGIHolmes} 上评估使用 SD-v1.4 训练的检测器能否泛化到近期生成器。
在 AIGI-Now 上，PE-SPC 以 98.0\% 的均值排名第一，PE-Linear 和 DINOv3-Linear 分别为 94.6\% 和 85.6\%（表~\ref{tab:aigi-now}）。
PE 的增益来自通过退化处理削弱低层生成痕迹的 sem 划分。
SPC 在全部九个 sem 子集上都提升了 PE 的性能，使均值从 91.3\% 升至 98.3\%，而 pix 均值变化很小，从 98.0\% 变为 97.8\%。
这表明，当低层线索被削弱时，SPC 有助于 PE 区分生成图像。
}

AIGI-Holmes includes generators with architectures different from that of SD-v1.4, including autoregressive models such as LlamaGen and VAR and diffusion transformers such as FLUX and SD3.5-L.
As shown in Table~\ref{tab:aigi-holmes}, SPC raises PE's mean from 97.6\% to 99.2\%.
Its three largest gains occur on Janus and the two Janus-Pro variants, with the largest improvement on Janus itself (91.7\% to 98.5\%).
MoEViE-Linear already reaches 99.4\%, and SPC improves its mean slightly to 99.5\%, the highest among the compared methods.
The two benchmarks show that PE's gains from SPC extend to generators beyond its SD-v1.4 detection training source.
\cn{
AIGI-Holmes 涵盖多种架构不同于 SD-v1.4 的生成器，包括 LlamaGen、VAR 等自回归模型，以及 FLUX、SD3.5-L 等扩散 Transformer。
如表~\ref{tab:aigi-holmes} 所示，SPC 将 PE 的均值从 97.6\% 提升至 99.2\%。
增益最大的三个子集均来自 Janus 系列，其中 Janus 本身的提升最明显，从 91.7\% 升至 98.5\%。
MoEViE-Linear 的均值已达到 99.4\%，SPC 进一步将其小幅提升至 99.5\%，为参与比较的方法中最高。
两个基准的结果说明 SPC 对 PE 的提升能够从检测训练源 SD-v1.4 迁移到其他生成器。
}

\end{document}